\documentclass[10pt]{njulink}

\usepackage{amsmath,amsfonts,bm}

\def\eqref#1{equation~\ref{#1}}

\def\1{\bm{1}}

\DeclareMathAlphabet{\mathsfit}{\encodingdefault}{\sfdefault}{m}{sl}
\SetMathAlphabet{\mathsfit}{bold}{\encodingdefault}{\sfdefault}{bx}{n}

\usepackage{graphicx}
\usepackage{booktabs}
\usepackage{tabularx}
\usepackage{array}
\usepackage[table]{xcolor}
\usepackage{pifont}
\usepackage{hyperref}
\usepackage{url}
\DeclareRobustCommand{\faLock}{\tikz[x=1em,y=1em,baseline=-0.08em]{%
  \draw[line width=0.09em] (0.18,0.42) -- (0.18,0.65)
    arc[start angle=180,end angle=0,radius=0.20] -- (0.58,0.42);
  \fill[rounded corners=0.04em] (0.08,0) rectangle (0.68,0.46);%
}}
\usepackage{tikz}
\usepackage{enumitem}
\usepackage{float}          
\usepackage{caption}        
\usepackage{makecell}       
\usepackage{multirow}
\usepackage{xcolor}
\usepackage{colortbl}
\usepackage{adjustbox}

\usepackage{wrapfig}
\usepackage{booktabs,array,makecell,tabularx}
\usepackage{xcolor,colortbl,pifont}

\definecolor{LWHeader}{HTML}{F4F5F7}
\definecolor{LWOurs}{HTML}{F0EBF8}
\definecolor{LWAccent}{HTML}{56416F}
\definecolor{LWNeutral}{HTML}{EEF0F3}

\definecolor{LWAction}{HTML}{86C9F4}  
\definecolor{LWMemory}{HTML}{9EDAA5}    
\definecolor{LWPhysics}{HTML}{F8A9BC}   
\definecolor{LWCausal}{HTML}{F0C66A}    
\definecolor{LWEvolution}{HTML}{BCA5EB} 
\definecolor{LWRevisit}{HTML}{FFC58F}      

\newcommand{\LWyes}{\textcolor{LWAccent}{\ding{51}}}
\newcommand{\LWno}{\textcolor{black!55}{\ding{55}}}
\newcommand{\cmark}{\LWyes}
\newcommand{\xmark}{\LWno}

\newcommand{\LWlock}{%
    \raisebox{0.05ex}{\scalebox{0.78}{\faLock}}%
}
\newcommand{\LWstats}[2]{#1\,/\,#2}

\newcommand{\LWtag}[2]{%
  \tikz[baseline=(LWtag.base)]{%
    \node[
      rounded corners=2pt,
      fill=#1!25,
      draw=#1,
      text=#1,
      line width=0.4pt,
      inner xsep=3pt,
      inner ysep=1.8pt,
      outer sep=0pt,
      font=\fontsize{7}{8}\selectfont\sffamily\bfseries
    ] (LWtag) {#2};%
  }%
}

\newcommand{\LWgap}{\hspace{1.2pt}}

\newcommand{\LWA}{\LWtag{LWAction}{A}}
\newcommand{\LWM}{\LWtag{LWMemory}{M}}
\newcommand{\LWP}{\LWtag{LWPhysics}{P}}

\newcommand{\LWE}{\LWtag{LWEvolution}{E}}
\newcommand{\LWS}{\LWtag{LWRevisit}{S}}

\usepackage{tcolorbox}
\tcbuselibrary{most}
\usepackage{xcolor}

\definecolor{PromptNavy}{HTML}{355C7D}
\definecolor{PromptPurple}{HTML}{8064A2}
\definecolor{PromptBlueBG}{HTML}{F4F8FB}
\definecolor{PromptPurpleBG}{HTML}{F8F5FB}

\newtcolorbox{rewritepromptbox}[1][]{
  enhanced,
  breakable,
  colback=PromptBlueBG,
  colframe=PromptNavy,
  coltitle=black,
  fonttitle=\bfseries\small,
  title={#1},
  boxrule=0.6pt,
  arc=2pt,
  left=7pt,
  right=7pt,
  top=6pt,
  bottom=6pt,
  before skip=6pt,
  after skip=8pt,
  fontupper=\footnotesize
}

\newtcolorbox{verifypromptbox}[1][]{
  enhanced,
  breakable,
  colback=PromptPurpleBG,
  colframe=PromptPurple,
  coltitle=black,
  fonttitle=\bfseries\small,
  title={#1},
  boxrule=0.6pt,
  arc=2pt,
  left=7pt,
  right=7pt,
  top=6pt,
  bottom=6pt,
  before skip=6pt,
  after skip=8pt,
  fontupper=\footnotesize
}

\njulinksetup{
  short-title={LongWM-Bench},
  type={},
}

\title{LongWM-Bench: Benchmarking Persistent and Navigable Worlds over Long-Horizon Interaction}
\author[1,*]{Jiaming Wang}
\author[1,*]{Yufei Chen}
\author[1,*]{Jiangtao Wu}
\author[2]{Zhiqi Bai}
\author[1]{Ziming Pan}
\author[1]{Yanghai Wang}
\author[4]{Fangqi~Zhu}
\author[2]{Chengting Yu}
\author[2]{Liangyu Yuan}
\author[2]{Qinghe Wang}
\author[2]{Jiaxin Xie}
\author[2]{Wenbo Su}
\author[3]{Zhaoxiang~Zhang}
\author[1,\dagger]{Jiaheng Liu}
\affiliation[1]{NJU}
\affiliation[2]{Alibaba ATH}
\affiliation[3]{CASIA}
\affiliation[4]{HKUST}
\contribution[*]{Equal contribution}
\contribution[\dagger]{Corresponding Author}
\correspondence{\email{jiaming_wang@smail.nju.edu.cn} \quad \email{liujiaheng@nju.edu.cn}}

\newcommand{\githuburl}{https://github.com/NJU-LINK/LongWM-Bench}

\metadata{\normalsize\textcolor{black}{\faGithub}\enspace\resourceurl{\githuburl}%
  }

\abstract{%
Short-term visual quality and action following offer limited evidence of whether an interactive world model remains reliable during extended use. We introduce LongWM-Bench, a benchmark of 300 tasks with target rollout durations of 60–180 seconds. Two task suites evaluate sustained control and revisit memory through combined inputs, repeated changes in control, and six families of trajectories with delayed and repeated revisits. We combine automatic visual and motion measurements with checklists tailored to each scene to assess visual quality, temporal consistency, revisit retention, physical and causal validity, and camera responses to actions. Evaluating 12 interactive world models reveals distinct capability profiles: models with similar appearance consistency differ substantially in revisit retention, while accurate camera responses can coexist with weak physical and causal validity. Coherent scene evolution remains a common weakness in our evaluation. These findings highlight the need to evaluate whether generated worlds remain controllable, retain previously observed content, and evolve coherently throughout interaction. 
}

\begin{document}
\maketitle
\enlargethispage{18pt}

\begin{figure}[H]
    \centering
    \vspace{10pt}
    \includegraphics[width=\linewidth]{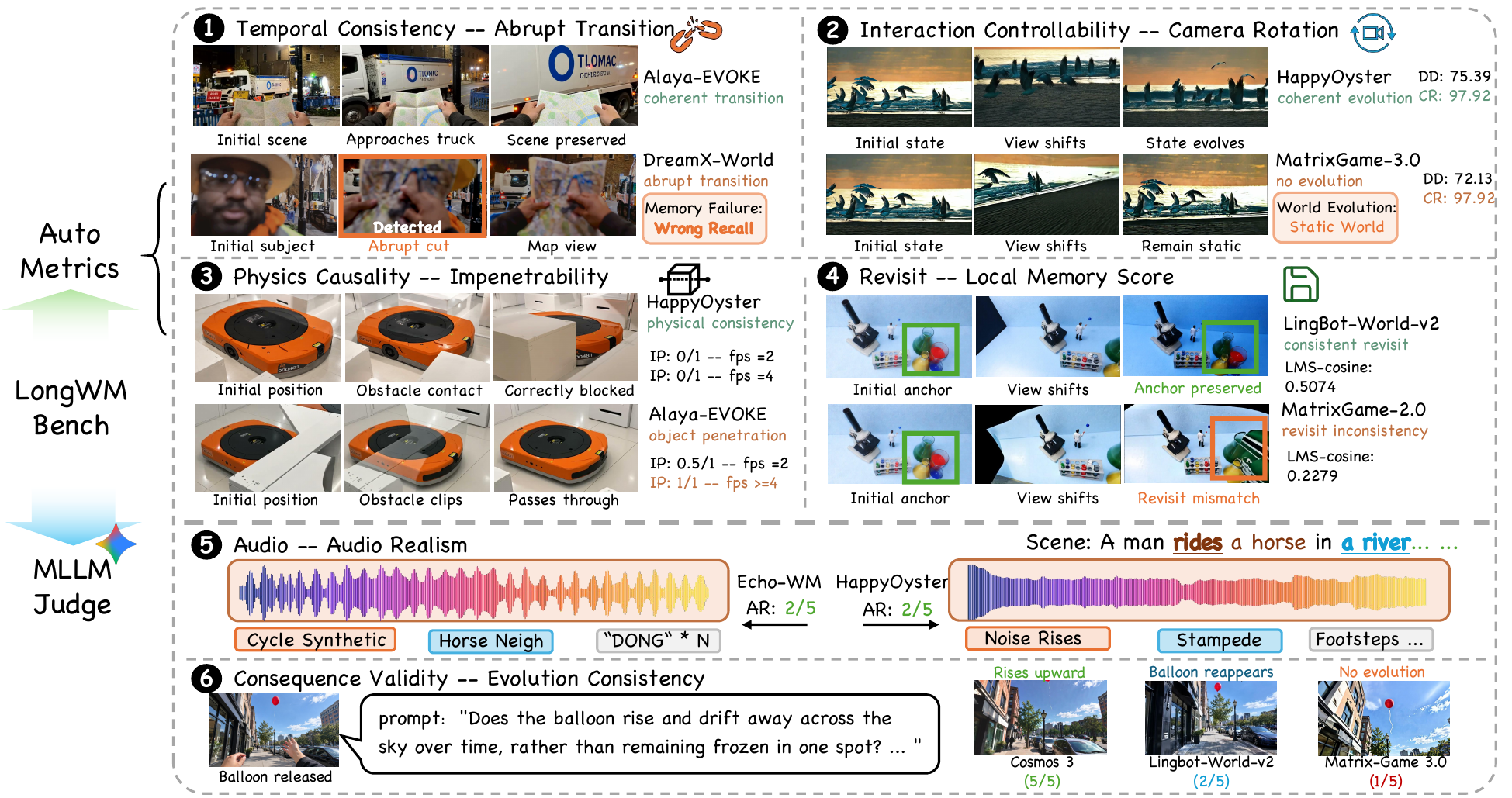}
    \vspace{-15pt}
    \caption{
   Overview of LongWM-Bench.
    }
    
    \label{fig:intro}
\end{figure}
\par\vspace{-3pt}
\noindent\begin{minipage}{\linewidth}
  \setlength{\parskip}{0pt}
  \raggedright\normalfont\footnotesize\color{black}
  \hrule width 0.22\linewidth height 0.4pt
  \vspace{3pt}
  \contributionlist\par
\end{minipage}\par
\providecommand{\MainContentsDetail}[1]{}
\clearpage
\begingroup
  \setcounter{tocdepth}{2}
  \renewcommand{\MainContentsDetail}[1]{\setcounter{tocdepth}{#1}}
  \setlength{\parskip}{0pt}
  \sffamily\fontsize{11.4}{15}\selectfont
  \color{LinkPurple}\hypersetup{linkcolor=LinkPurple,linktoc=all}
  \def\AppendixContentsStart{\par\addvspace{10pt}%
    {\large\bfseries\color{LinkPurple}Appendices\par}\nobreak
    \setcounter{tocdepth}{1}}
  \makeatletter
  \renewcommand{\l@section}[2]{\par\addvspace{4pt}%
    {\bfseries\@dottedtocline{1}{0em}{1.9em}{#1}{#2}}}
  \renewcommand{\l@subsection}{\@dottedtocline{2}{1.15em}{2.5em}}
  \renewcommand{\l@subsubsection}{\@dottedtocline{3}{2.8em}{3em}}
  \renewcommand{\l@paragraph}[2]{%
    {\fontsize{10.3}{13.5}\selectfont\@dottedtocline{4}{3.65em}{0em}{#1}{#2}}}
  \makeatother
  \section*{Contents}
  \InputIfFileExists{\jobname.toc}{}{}
  \setcounter{tocdepth}{2}
\endgroup
\clearpage

\section{Introduction}
\label{sec:introduction}
Interactive world models are moving beyond video generation toward
environments that users and agents can continuously explore and
interact with~\citep{ICLR2025_b71ecea2,
zhang2025matrixgameinteractiveworldfoundation,
mao2025yumeinteractiveworldgeneration}. Systems such as
HappyOyster already support real-time exploration and interaction
within generated worlds~\citep{happyoyster}. Yet sustaining
interaction does not by itself ensure that the generated world
remains reliable over time~\citep{wang2026matrixgame30realtimestreaming,
zhang2026mbenchcomprehensivebenchmarkmemory, bai2026happyworld}. A sequence of movements
and turns may appear correct locally, while scene layouts gradually
drift, object states change unexpectedly, or traces of earlier actions
disappear upon return. When generated observations guide subsequent actions, evaluation must consider whether the world remains coherent throughout interaction, beyond the quality of individual frames and immediate responses to actions.

Recent benchmarks have expanded interactive world-model evaluation
beyond visual quality to multi-turn control, scene consistency,
physics, and memory~\citep{xu2026worldmarkunifiedbenchmarksuite,
fang2026iworldbenchbenchmarkinteractiveworld}. WBench evaluates a
broad range of interactive-generation capabilities~\citep{wbench},
while WorldRoamBench examines action, vision, physics, and memory over
10--60-second interactions~\citep{worldroambench}. These efforts
provide an important foundation, but evaluations at such horizons
cannot determine whether reliability persists as interaction extends
from tens of seconds to several minutes. For sustained control, the
question is whether correct responses can be maintained under long
sequences of combined and changing actions. For revisit memory, the
question is whether previously established appearance, spatial
structure, and state information survive prolonged absence and
repeated returns. More generally, long-horizon evaluation should test
whether later observations and interactions remain grounded in the
world established earlier, rather than merely whether generation can
continue for longer.

We introduce \textbf{LongWM-Bench}, a benchmark for long-horizon
navigability comprising 300 tasks with 60--180-second rollouts.
The benchmark targets two complementary challenges of prolonged
interaction. \textbf{Complex control} tests whether models remain responsive under combined actions, direction reversals, and repeated changes in control.
\textbf{Revisit memory} tests whether previously observed scenes remain visually consistent after extended periods out of view and repeated returns. Together, the two suites assess sustained controllability and visual persistence during prolonged interaction.

Figure~\ref{fig:intro} provides an overview of our evaluation
framework. LongWM-Bench evaluates interactive world models along
five complementary capability dimensions: \textbf{Action
Controllability}, \textbf{Perceptual Quality}, \textbf{Physical \&
Causal Validity}, \textbf{Temporal Consistency}, and
\textbf{Revisit Memory}. We combine visual and motion measurements,
revisit-based comparisons, and event-level checks to assess these
capabilities throughout each rollout. Importantly, consistency does
not require the world to remain static: action-induced changes should
persist unless later events justify otherwise, while ongoing processes
should continue to evolve coherently. This enables us to diagnose a
central failure mode of current interactive world models: a model may
respond correctly to individual actions while gradually drifting away
from the world established earlier. Our evaluation therefore examines
not only whether local responses are correct, but also whether later
states remain consistent with prior observations, actions, and their
consequences.

Our contributions are threefold:

\begin{itemize}[leftmargin=1.5em]
\item \textbf{\textit{LongWM-Bench.}}
We introduce a benchmark for long-horizon navigability with
300 tasks and 60--180-second rollouts. It targets two complementary
challenges---sustained control and world persistence---through
complex action sequences, delayed revisits, and repeated revisits.

    \item \textbf{\textit{Diagnostic evaluation.}}
    We develop a multidimensional evaluation framework that combines
    automatic visual and motion measurements, revisit-based comparisons,
    and MLLM-based event-level judgments. It evaluates five complementary
    capabilities---action controllability, perceptual quality, physical
    and causal validity, temporal consistency, and revisit memory---
    using observable evidence accumulated throughout each rollout.

    \item \textbf{\textit{Long-horizon analysis.}}
    We evaluate 12 interactive world models and analyze their capabilities during extended interaction. The results reveal distinct weaknesses in camera control, revisit retention, physical and causal validity, and ongoing scene evolution, showing why these capabilities should be evaluated together.
\end{itemize}

\begin{table}[H]
\centering
\caption[
Comparison with interactive world model benchmarks.
]{
Comparison of LongWM-Bench with representative world
model benchmarks. Revisit denotes the number of revisit categories
and unique trajectories; Turns is the average number of interaction
turns per instance, and Duration is measured in seconds.
Capabilities include \protect\LWA\ (action control),
\protect\LWS\ (single revisit), \protect\LWM\ (multiple revisits),
\protect\LWP\ (physics), and \protect\LWE\ (ongoing evolution).
Audio indicates audio support, and a lock marks unavailable
information for non-open-source benchmarks.
}
\label{tab:benchmark_comparison}


\begingroup
\fontsize{8pt}{8.5pt}\selectfont
\renewcommand{\arraystretch}{1.18}
\setlength{\tabcolsep}{3pt}
\setlength{\aboverulesep}{0pt}
\setlength{\belowrulesep}{0pt}

\resizebox{\linewidth}{!}{%
\begin{tabular}{@{}lccccccc@{}}
\toprule[0.85pt]

\rowcolor{LWHeader}
\textbf{Benchmark}
& \textbf{Revisit}
& \textbf{Instances}
& \textbf{Domains}
& \textbf{Turns}
& \textbf{Duration}
& \textbf{Audio}
& \textbf{Capabilities}
\\

\midrule[0.45pt]

WorldScore~\citep{Duan_2025_ICCV}
& --
& 3,000
& 15
& 1.3
& 2--10
& \xmark
& \LWA\LWgap\LWE
\\

WorldMark~\citep{xu2026worldmarkunifiedbenchmarksuite}
& \LWstats{2}{6}
& 500
& 3
& 1.7
& 20/40/60
& \xmark
& \LWA\LWgap\LWS
\\

iWorld-Bench~\citep{fang2026iworldbenchbenchmarkinteractiveworld}
& \LWstats{1}{8}
& 4,900
& 2
& 1.0
& $\sim$ 5
& \xmark
& \LWA\LWgap\LWS
\\

WBench~\citep{wbench}
& \LWstats{2}{23}
& 289
& 6
& 3.7
& 10--45
& \xmark
& \LWA\LWgap\LWS\LWgap\LWP
\\

WorldOlympiad~\citep{zhao2026worldolympiadworldmodelsurvive}
& --
& 1,000
& 3
& 3.2
& $2$--$60$
& \xmark
& \LWA\LWgap\LWP\LWgap\LWE
\\

WorldRoamBench~\citep{worldroambench}
& \LWlock
& \LWlock
& \LWlock
& \LWlock
& 10--60
& \xmark
& \LWA\LWgap\LWS\LWgap\LWP
\\

\midrule[0.45pt]

\rowcolor{LWOurs}
\textbf{LongWM-Bench}
& \LWstats{\textbf{6}}{\textbf{150}}
& \textbf{300}
& \textbf{9}
& \textbf{40.8}
& \textbf{60--180}
& \cmark
& \LWA\LWgap\LWS\LWgap\LWM\LWgap\LWP\LWgap\LWE
\\

\bottomrule[0.85pt]
\end{tabular}%
}

\endgroup
\end{table}

\section{Related Work}

\paragraph{Interactive World Models}
\label{sec:related_models}

Interactive world models extend video generation into environments
that respond to actions. Genie learns latent controls from videos
without action labels~\citep{bruce2024geniegenerativeinteractiveenvironments},
while GameNGen demonstrates diffusion-based real-time game
simulation~\citep{ICLR2025_b71ecea2}. Matrix-Game 2.0 and
Hunyuan-GameCraft advance continuous interaction through keyboard
and mouse control, history conditioning, and efficient
inference~\citep{he2025matrix,li2025hunyuan}. Genie 3 and
LingBot-World support minute-scale interaction across diverse
scenes~\citep{team2026advancing}, while SANA-WM improves efficiency
with hybrid linear attention and dual-branch camera
control~\citep{zhu2026sanawmefficientminutescaleworld}.
HappyOyster broadens user intervention through its Wandering and
Directing modes, supporting exploration and real-time editing of
characters, cameras, and scenes~\citep{happyoyster}.
Echo-WM unifies discrete commands and continuous poses as relative
6-DoF trajectories, enabling first- and third-person navigation
with joint audio--video
generation~\citep{zhang2026echowmopenenterableomnimodal}.
These advances improve efficiency, control, and interaction
length, but do not by themselves establish whether the generated
world remains consistent.

\paragraph{World Model Evaluation}
\label{sec:related_benchmarks}

Table~\ref{tab:benchmark_comparison} compares LongWM-Bench with
representative world-model evaluation benchmarks. Early video-generation
benchmarks such as VBench and EvalCrafter focus on visual quality,
motion, temporal consistency, and text alignment~\citep{vbench,Liu_2024_CVPR, wu2025vidic, wen2026ave},
while VBench-2.0 further introduces physical and commonsense
evaluation~\citep{zheng2025vbench20advancingvideogeneration}.
Recent benchmarks increasingly target interactive settings:
WorldScore evaluates controllability, quality, and dynamics under
specified camera trajectories;
WorldMark and iWorld-Bench examine action responses, view consistency,
control, and memory~\citep{xu2026worldmarkunifiedbenchmarksuite,fang2026iworldbenchbenchmarkinteractiveworld};
and WBench and WorldRoamBench extend evaluation to multi-turn interaction,
physics, and memory~\citep{wbench,worldroambench}. PlayWorld further
introduces an agent that adapts actions to visual feedback~\citep{playworld}.
Together, these benchmarks provide broad diagnostics, but long-horizon
reliability remains less characterized. Existing evaluations typically
focus on interaction horizons of around 60 seconds or shorter, leaving
open whether controllability, world consistency, and memory persist over
longer interactions and repeated revisits. LongWM-Bench complements these
efforts by evaluating sustained controllability and world persistence over
60--180-second rollouts.

\section{LongWM-Bench}
\label{sec:longwm_compass}

\subsection{Dataset Construction}
\label{sec:data_construction}

\begin{figure}[tb]
    \centering
    \includegraphics[width=\linewidth]
    {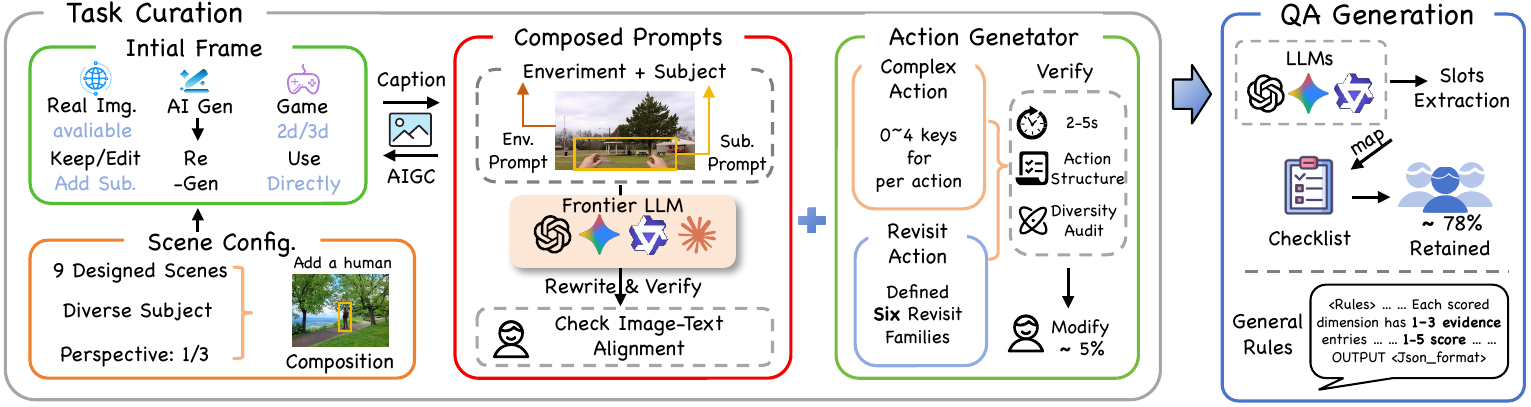}
    \caption[Data curation pipeline of LongWM-Bench.]{
Data curation pipeline of LongWM-Bench.
The pipeline comprises three stages: task curation with prompt rewriting and verification, interaction-sequence generation and validation, and checklist-based QA generation with human review.
    }
    \label{fig:pipeline}
\end{figure}

Figure~\ref{fig:pipeline} summarizes our three-stage data construction
pipeline, which combines multi-model generation, automated verification,
and human review to construct diverse interaction trajectories and
reliable evaluation checklists. All prompts are provided in
Appendix~\ref{app:prompt}.

\textbf{Stage 1: Initial-frame and caption construction.}
We curate initial frames from real images, game screenshots, and
GPT-Image-2 generations across nine scene domains
~\citep{openai2026gptimage2}.
Gemini 3.8 Flash first produces \textit{environment},
\textit{subject}, and \textit{perspective} descriptions
~\citep{google2026gemini38flash}.
The descriptions are then independently rewritten and verified using
GPT-5.6 Sol, Qwen3.8-Max, Gemini 3.8 Flash, and Claude Sonnet 5
~\citep{openai2026gpt56sol,alibaba2026qwen38max,
google2026gemini38flash,anthropic2026claudesonnet5}.
Human reviewers check image--text alignment and correct 10\% of the
samples. The resulting image--text pairs provide the scene
specification used as input for interaction generation.

\textbf{Stage 2: Interaction construction.}
Using the same four-model pool, we generate complex-control sequences
and six predefined revisit families to improve trajectory diversity.
Generated trajectories are automatically checked for action structure,
duration, revisit validity, and diversity, followed by human revision
of invalid, repetitive, or degenerate sequences; approximately 5\%
are modified.

\textbf{Stage 3: Checklist construction.}
GPT-5.6 Sol, Qwen3.8-Max, and Gemini 3.8 Flash extract
capability-specific evaluation slots from each scene and trajectory
and convert them into candidate checklist items
~\citep{openai2026gpt56sol,alibaba2026qwen38max,
google2026gemini38flash}.
Human reviewers remove redundant, speculative, ambiguous, or visually
unverifiable items, while ensuring that the retained checks are grounded
in observable evidence and relevant to the target capability;
approximately 78\% of the candidates are retained. The final per-sample
checklists are supplemented with shared evidence requirements and
1--5 scoring rules for consistent evaluation across models.

\subsection{Data Statistics}
\begin{figure}[H]
    \centering
    \includegraphics[width=0.95\linewidth]
    {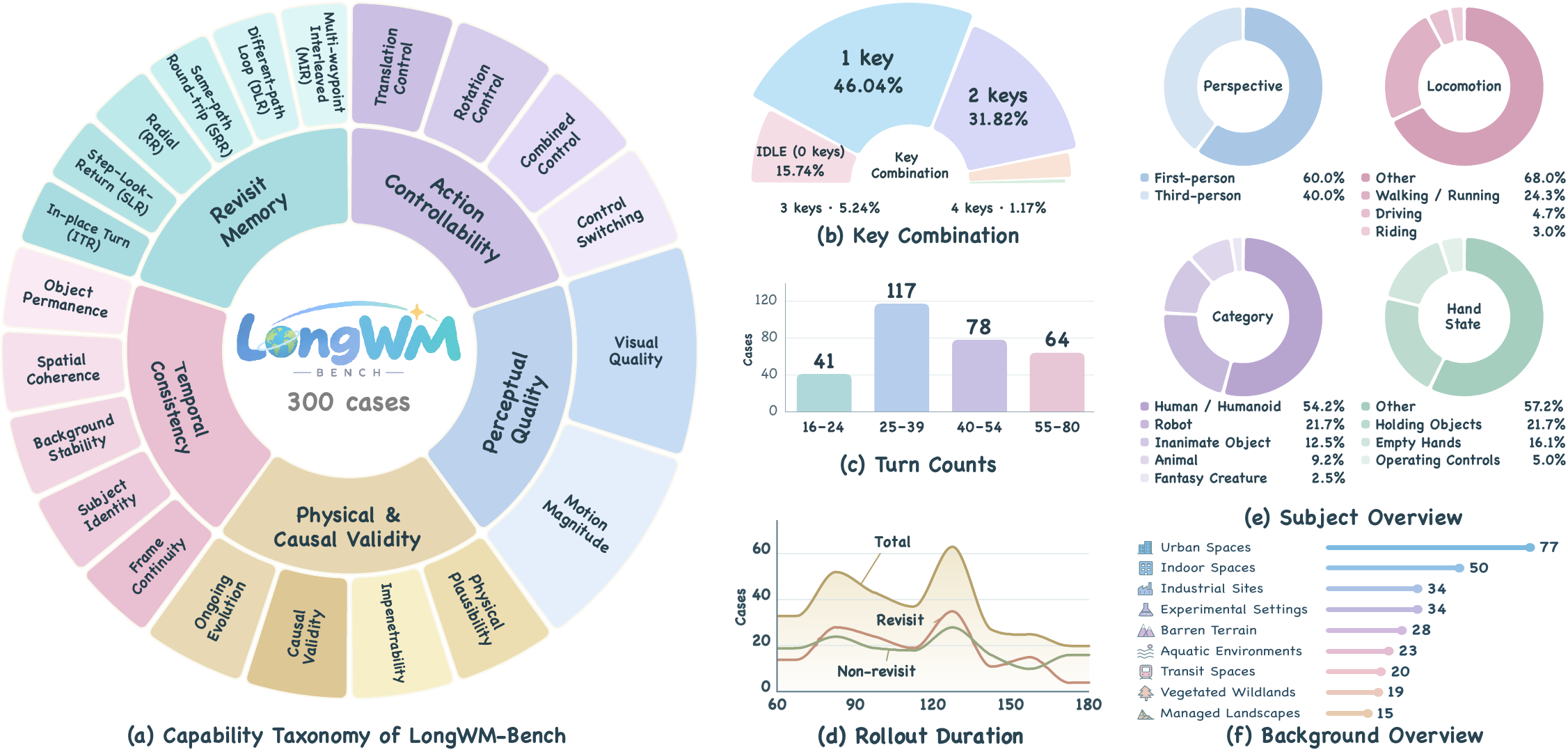}
    \caption{
    Statistics of LongWM-Bench, including capability coverage,
    interaction patterns, rollout duration, background domains,
    and subject distributions.
    }
    \label{fig:data_statistics}
\end{figure}
LongWM-Bench contains 150 complex-control cases and 150 revisit
cases, with requested durations of 60--180 seconds.
Figure~\ref{fig:data_statistics} summarizes the distribution
of scenes, viewpoints, subjects, and control programs.
First- and third-person perspectives account for 60\% and 40\%
of the cases, respectively. Across all annotated action
segments, including idle intervals, 38.2\% contain simultaneous
key inputs. The six revisit families range from in-place turns
and short out-and-back movements to radial excursions,
closed loops, and interleaved waypoint visits.
These patterns distinguish repeated observation of one anchor
from returns after longer or multi-location excursions;
the detailed trajectory definitions and examples are given in Appendix~\ref{app:revisit_suite}.
Across the 300 cases, the final checklists contain 1,405 items (4.68 per case).
Of these, 736 assess PC (52.4\%), 314 assess CC (22.3\%), and 355 assess EC (25.3\%).
Each case includes the dimensions applicable to its scene.

\subsection{Evaluation Metrics}
\addtocontents{toc}{\protect\MainContentsDetail{4}}
\label{sec:evaluation_metrics}

Building on the capability taxonomy in Figure~\ref{fig:data_statistics},
we evaluate interactive world models along five complementary dimensions:
\textbf{Perceptual Quality}, \textbf{Temporal Consistency},
\textbf{Revisit Memory}, \textbf{Physical \& Causal Validity}, and
\textbf{Action Controllability}. Together, these dimensions cover both
the local quality of generated observations and the long-horizon
reliability of the evolving world under sustained interaction. The
taxonomy defines the capabilities expected of a persistent and navigable
world, while the evaluation metrics operationalize these capabilities
through observable evidence in generated rollouts. Our evaluation combines
perceptual, motion, visual-feature, geometric, and vision--language
measurements and reports each metric separately to preserve diagnostic
interpretability. Detailed definitions and evaluator configurations are provided in Appendix~\ref{sec:video_metrics}.

\paragraph{Perceptual Quality.}
This dimension characterizes whether generated observations remain
visually plausible while exhibiting meaningful visual dynamics.
Following the capability taxonomy, we assess both visual quality and
motion magnitude.
\textbf{Visual Quality (VQ)} uses MUSIQ
~\citep{ke2021musiqmultiscaleimagequality} to score individual frames
and averages the frame-level predictions, providing a reference-free
measure of visual fidelity throughout the rollout.
\textbf{Dynamic Degree (DD)} estimates motion magnitude from dense
optical flow, with normalization for temporal sampling and image scale
and trimming to reduce the influence of extreme flow estimates.
Reporting VQ and DD together distinguishes visually convincing but
nearly static generations from outputs with sustained visual dynamics,
while DD itself reflects motion magnitude rather than physical
correctness. 

\paragraph{Temporal Consistency.}
This dimension evaluates whether the visible world remains coherent
throughout continuous interaction, including observation continuity
and the temporal stability of both subjects and backgrounds.
\textbf{Jump-Based Continuity (JP)} uses
TransNetV2~\citep{soucek2024transnet} to detect abrupt cut-like
transitions, with the strongest predicted cut probability used to
measure temporal continuity.
\textbf{Subject Consistency (SC)} measures the stability of the
foreground subject by combining adjacent-frame DINOv2 similarity with
CLIP similarity to the initial observation, capturing both short-term
appearance consistency and accumulated subject drift.
\textbf{Background Consistency (BC)} separately measures the stability
of the background using adjacent-frame and initial-observation
comparisons over background regions.
Subject masks are used where available to separate foreground subjects
from background content.

\paragraph{Revisit Memory.}
This dimension evaluates whether previously observed locations remain
consistent after leaving the field of view and being revisited later.
We evaluate 150 predefined revisit trajectories with designated
observation anchors.
\textbf{Global Memory Score (GMS)} uses DINOv2
features~\citep{oquab2024dinov2learningrobustvisual} to compare each
revisit with the anchor's initial observation, measuring retention of
the originally observed scene.
\textbf{Local Memory Score (LMS)} instead compares consecutive revisits,
excluding the initial-to-first-revisit pair, and therefore applies to
trajectories with at least three observations.
Both scores weight comparisons by the elapsed time since the initial
observation, emphasizing consistency over longer intervals.
GMS captures long-term consistency with the original view, whereas LMS
captures stability across successive returns. Appendix~\ref{app:pose_conditioning} reports rankings on pose-qualified revisits and shared observation pairs.

\paragraph{Physical \& Causal Validity.}
This dimension evaluates whether the generated world follows
scene-appropriate physical constraints, causal relationships, and
ongoing state evolution.
\textbf{Physical Causality (PC)} assesses physical plausibility through
properties and responses such as gravity, support, buoyancy, collision,
deformation, material behavior, thermodynamics, and non-penetration of solid boundaries.
\textbf{Content Causality (CC)} evaluates whether a scene-supported
cause produces a visually consistent effect on another object or region,
such as illumination inducing corresponding shadows or reflections.
\textbf{Evolution Consistency (EC)} measures whether spontaneous
processes and state changes, such as drifting, flowing, or melting,
evolve coherently over time rather than remaining unnaturally static.
PC, CC, and EC are evaluated by Gemini 3.1 Pro Preview~\citep{google2026gemini31pro} using
case-specific checklists, with each item scored
on a 1--5 scale based on visual evidence.

\paragraph{Action Controllability.}
This dimension evaluates whether camera motion follows each
model's native control inputs (camera trajectories or actions)
in first- and third-person settings during translation, rotation,
combined controls, and repeated control switching.
Using camera trajectories recovered by
ViPE~\citep{huang2025vipevideoposeengine},
\textbf{Translation Response (TR)} measures whether the net camera
displacement of each action segment agrees with the commanded
translation direction in the action segment's initial camera coordinate
system.
\textbf{Camera Rotation Response (CR)} measures whether the estimated
yaw and pitch changes follow the commanded rotation directions with
sufficient angular magnitude.
Both metrics report the proportion of correctly executed action
segments, with combined translation--rotation commands contributing to
both TR and CR.
TR and CR measure directional camera response. Appendix~\ref{app:control_alignment}
describes how each model receives the prescribed inputs.

Together, these metrics provide complementary views of long-horizon
reliability. Perceptual and Temporal metrics characterize the
quality and continuity of the generated observations, while revisit,
physical--causal, and controllability metrics test whether the world
preserves previously established states, evolves plausibly, and remains
responsive to sustained interaction. We therefore report these metrics separately throughout our experiments
to facilitate fine-grained diagnostic analysis across complementary
long-horizon capabilities.

\section{Experiments}
\addtocontents{toc}{\protect\MainContentsDetail{2}}
\label{sec:experiments}
\begin{table}[H]
    \centering
    \caption{
        Results on LongWM-Bench across five evaluation dimensions.
        Closed-source models are listed first, followed by open-weight models;
        all models are compared jointly for each metric (higher is better).
        The highest score in each metric is \textbf{bold} with a
        category-matched pale background; the second highest is
        \underline{underlined}.
        Params. denotes the parameter count;
        \raisebox{0.1ex}{\scriptsize\faLock} indicates closed-source.
        All scores are normalized, higher is better. 
    }
    \label{tab:main_results}

    \begingroup

    \newcommand{\wmBestCell}[2]{%
      \begingroup
      \setlength{\fboxsep}{0pt}%
      \makebox[\linewidth][c]{%
        \colorbox{#1}{%
          \makebox[\dimexpr\linewidth-0.2pt\relax][c]{%
            \strut\textbf{#2}}}}%
      \endgroup
    }
    
    \newcommand{\wmBestP}[1]{\wmBestCell{wmBestPerception}{#1}}
    \newcommand{\wmBestT}[1]{\wmBestCell{wmBestTemporal}{#1}}
    \newcommand{\wmBestR}[1]{\wmBestCell{wmBestRevisit}{#1}}
    \newcommand{\wmBestC}[1]{\wmBestCell{wmBestCausality}{#1}}
    \newcommand{\wmBestI}[1]{\wmBestCell{wmBestInteraction}{#1}}
    \newcommand{\wmSecond}[1]{\underline{#1}}

    \definecolor{wmPerception}{HTML}{E8F1FA}
    \definecolor{wmTemporal}{HTML}{EEEAF6}
    \definecolor{wmCausality}{HTML}{FBEEE2}
    \definecolor{wmInteraction}{HTML}{E7F3EA}

    \definecolor{wmRevisit}{HTML}{FBEAF1}

    \definecolor{wmBestPerception}{HTML}{CEE5F9}
    \definecolor{wmBestTemporal}{HTML}{E2D9F4}
    \definecolor{wmBestCausality}{HTML}{F8DEC5}
    \definecolor{wmBestInteraction}{HTML}{D4EBD9}

    \definecolor{wmBestRevisit}{HTML}{F4D2E1}

    \setlength{\tabcolsep}{1.6pt}
    \renewcommand{\arraystretch}{1.20}
    \fontsize{8.0}{9.0}\selectfont

    \newlength{\metriccolwidth}
    \setlength{\metriccolwidth}{0.79cm}

    \newcommand{\wmHeaderBox}[3]{%
        \begingroup
        \setlength{\fboxsep}{0pt}%
        \colorbox{#1}{%
            \parbox[c][18pt][c]{#2}{%
                \centering
                #3\par
            }}%
        \endgroup
    }

    \begin{adjustbox}{max width=\textwidth,center}
    \begin{tabular}{
        @{}l c *{12}{>{\centering\arraybackslash}p{\metriccolwidth}}@{}
    }
    \toprule

    \multirow{2}{*}{\textbf{Model}}
    & \multirow{2}{*}{\textbf{Params.}}
        & \multicolumn{2}{c}{%
        \wmHeaderBox{wmPerception}
        {\dimexpr2\metriccolwidth+2\tabcolsep+4pt\relax}
        {\textbf{Perceptual}\\\textbf{Quality}}}

    & \multicolumn{3}{c}{%
        \wmHeaderBox{wmTemporal}
        {\dimexpr3\metriccolwidth+4\tabcolsep+4pt\relax}
        {\textbf{Temporal}\\\textbf{Consistency}}}

    & \multicolumn{2}{c}{%
        \wmHeaderBox{wmRevisit}
        {\dimexpr2\metriccolwidth+2\tabcolsep+4pt\relax}
        {\textbf{Revisit}\\\textbf{Memory}}}

    & \multicolumn{3}{c}{%
        \wmHeaderBox{wmCausality}
        {\dimexpr3\metriccolwidth+4\tabcolsep+4pt\relax}
        {\textbf{Physical \& Causal}\\\textbf{Validity}}}

    & \multicolumn{2}{c}{%
        \wmHeaderBox{wmInteraction}
        {\dimexpr2\metriccolwidth+2\tabcolsep+4pt\relax}
        {\textbf{Action}\\{\fontsize{7.5}{9}\selectfont\textbf{Controllability}}}}
    \\
        
    \cmidrule(lr){3-4}
    \cmidrule(lr){5-7}
    \cmidrule(lr){8-9}
    \cmidrule(lr){10-12}
    \cmidrule(l){13-14}

    &
    & \textbf{VQ}
    & \textbf{DD}
    & \textbf{JP}
    & \textbf{SC}
    & \textbf{BC}
    & \textbf{GMS}
    & \textbf{LMS}
    & \textbf{PC}
    & \textbf{CC}
    & \textbf{EC}
    & \textbf{TR}
    & \textbf{CR}
    \\

        \midrule[0.72pt]

        \multicolumn{14}{@{}c@{}}{
            \textbf{Closed-source models}}
        \\
        \midrule[0.62pt]
        HappyOyster
        & \LWlock
        & 63.91
        & 69.77
        & 59.54
        & 76.75
        & \wmSecond{84.70}
        & 62.78
        & 62.50
        & 50.72
        & 47.84
        & 21.77
        & 79.64
        & 93.33
        \\
        Genie 3\textsuperscript{\textdagger}
        & \LWlock
        & 64.60
        & \wmBestP{87.81}
        & 79.95
        & 76.70
        & 82.57
        & N/A
        & N/A
        & \wmBestC{67.44}
        & 55.28
        & 21.55
        & N/A
        & N/A
        \\

        \midrule[0.72pt]
    
        \multicolumn{14}{@{}c@{}}{
            \textbf{Open-weight models}}
        \\
        \midrule[0.62pt]

        DreamX-World
        & 5B
        & 62.89
        & 58.28
        & 85.39
        & 77.59
        & 84.02
        & 68.92
        & 72.25
        & 46.11
        & \wmSecond{56.25}
        & \wmBestC{35.70}
        & \wmBestI{94.26}
        & 90.32
        \\

        Lyra 2.0
        & 14B
        & 66.90
        & 55.83
        & 79.71
        & 79.39
        & 84.04
        & \wmBestR{79.81}
        & \wmBestR{86.78}
        & 36.62
        & 36.38
        & 12.86
        & 69.03
        & \wmBestI{95.88}
        \\

        Echo-WM
        & 19B
        & 65.99
        & 62.61
        & 79.79
        & 78.86
        & 84.31
        & 56.80
        & 62.61
        & 36.59
        & 37.50
        & 17.96
        & 84.49
        & 92.89
        \\

        Alaya-EVOKE
        & 14B
        & \wmBestP{68.61}
        & 51.42
        & \wmSecond{86.25}
        & 78.00
        & 84.44
        & 69.46
        & 77.33
        & \wmSecond{63.11}
        & \wmBestC{60.79}
        & \wmSecond{29.67}
        & 59.50
        & 42.67
        \\
        
        LingBot-World-v2
        & 14B
        & 60.43
        & 55.01
        & 73.38
        & 76.32
        & 81.54
        & 46.41
        & 52.72
        & 44.13
        & 32.59
        & 14.15
        & \wmSecond{85.38}
        & 88.21
        \\

        Matrix-Game 3.0
        & 5B
        & 64.95
        & \wmSecond{70.99}
        & 57.18
        & 78.09
        & 80.48
        & 39.41
        & 51.02
        & 8.15
        & 4.54
        & 0.93
        & 81.76
        & \wmSecond{95.55}
        \\
        
        Cosmos~3
        & 16B
        & 66.06
        & 25.30
        & \wmBestT{88.87}
        & \wmBestT{82.19}
        & \wmBestT{86.91}
        & \wmSecond{79.27}
        & \wmSecond{82.57}
        & 36.30
        & 33.33
        & 10.56
        & 42.49
        & 23.64
        \\
    
        Yume-1.5
        & 5B
        & \wmSecond{68.06}
        & 49.36
        & 85.08
        & 79.02
        & 83.35
        & 47.89
        & 58.36
        & 24.32
        & 24.63
        & 12.28
        & 58.03
        & 24.44
        \\
    
        ABot-World
        & 0.5B
        & 50.34
        & 69.55
        & 57.78
        & 76.61
        & 77.64
        & 34.57
        & 41.33
        & 17.44
        & 11.16
        & 4.09
        & 60.36
        & 81.73
        \\

        Matrix-Game 2.0
        & 1.8B
        & 53.85
        & 52.72
        & 80.33
        & \wmSecond{79.43}
        & 79.00
        & 24.78
        & 34.65
        & 2.08
        & 0.89
        & 0.00
        & 79.94
        & 90.23
        \\

        \bottomrule
    \end{tabular}
    \end{adjustbox}
    
    \vspace{2pt}

    \begin{minipage}{\textwidth}
    \footnotesize
    \textsuperscript{\textdagger}\,Genie 3 is evaluated through its official hosted interface with a maximum rollout duration of 60 seconds;
    unavailable long-horizon metrics are denoted by ``N/A''.
    \end{minipage}
    
    \endgroup
    \vspace{-2mm}
\end{table}
\clearpage
\subsection{Experimental Setup}
\label{sec:experimental_setup}

We evaluate 12 interactive world models:
ABot-World~\citep{jiang2026abot},
Alaya-EVOKE~\citep{yin2026alaya},
DreamX-World~\citep{team2026dreamx},
Genie 3~\citep{genie3},
HappyOyster~\citep{happyoyster},
Echo-WM~\citep{zhang2026echowmopenenterableomnimodal},
LingBot-World-v2~\citep{lingbotworldv2},
Lyra 2.0~\citep{shen2026lyra},
Matrix-Game 2.0~\citep{he2025matrix},
Matrix-Game 3.0~\citep{wang2026matrixgame30realtimestreaming},
Cosmos~3~\citep{nvidia2026cosmos3},
and Yume-1.5~\citep{mao2026yume1}.
For each case, the initial input, task requirements, and interaction
budget are held fixed across models, with controls mapped to each
model's native interface.
Complex control and revisit memory use prescribed action sequences.
We evaluate the generated videos using the automatic measurements
and VLM-based checks defined in Section~\ref{sec:evaluation_metrics}.
Model details and
inference settings are in Appendix~\ref{app:video-inference}.


\subsection{Main Results}
\label{sec:main_results}

Table~\ref{tab:main_results} summarizes performance on LongWM-Bench.
We report visual quality, world consistency, physical and causal
validity, control accuracy, and revisit similarity as separate scores.
Metrics are computed wherever their required evidence is available,
including control and revisit measurements.
Scores are aggregated from case-level results, with per-suite
breakdowns and evaluation coverage provided in the appendix.


\begin{figure}[H]
    \centering
    \vspace{-10pt}
    \includegraphics[width=\linewidth]
    {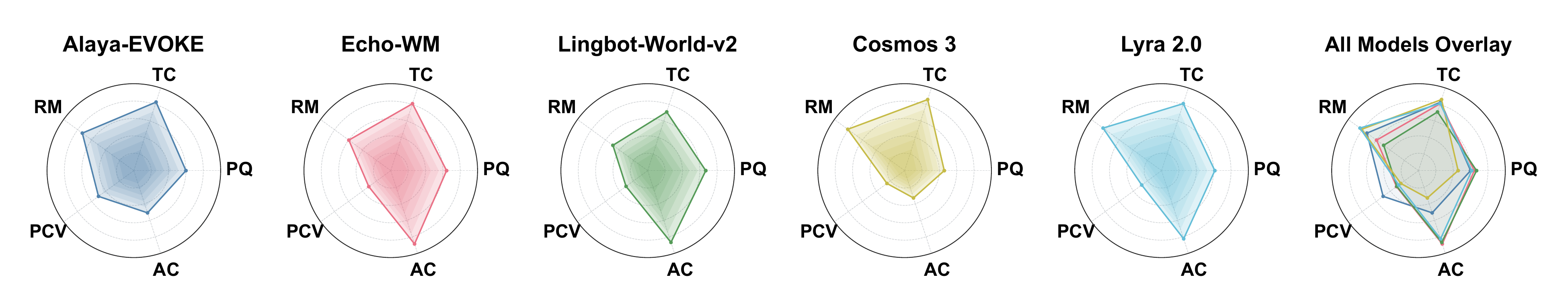}
    \vspace{-10pt}
    \caption{
     Multi-metric radar comparison on five evaluation dimensions. Each axis reports the average of the corresponding dimension's submetric scores.
    }
    \vspace{-5pt}
    \label{fig:radar-five}
\end{figure}

\paragraph{Uneven capabilities, with world evolution as a shared weakness.}
Table~\ref{tab:main_results} and Figure~\ref{fig:radar-five} show distinct capability profiles across the five dimensions.
Frame-level visual quality scores cluster in a relatively narrow range, with 10 of the 12 models scoring between 60 and 70 on VQ.
Models generally struggle when a scene process must continue to evolve. Eight models score below 20 on evolution consistency (EC), and the highest score is 35.70.
Alaya-EVOKE scores above 60 on both physical and content causality checks but reaches only 29.67 on EC.
Even when a model preserves physical and causal relationships between objects, it can struggle to sustain a change already underway.

\paragraph{Similar appearance consistency can mask large differences in revisit retention.}
When a scene leaves the field of view and is revisited, the model must retain its previously observed layout and appearance.
Lyra 2.0 and Matrix-Game 2.0 show nearly identical subject consistency across frames, both scoring about 79.4 on SC.
When the revisited scene is compared with its first observation, however, their global memory scores (GMS) are 79.81 and 24.78, respectively (Table~\ref{tab:main_results}).
The reference observation also matters across models. For 10 of the 11 models with revisit scores, the scene more closely resembles the preceding revisit (LMS) than the first observation (GMS).

\paragraph{Strong control does not ensure a reliable world.}
Navigation requires the camera to turn as instructed and the scene to retain its appearance when the camera turns back.
Cosmos~3 preserves the initially observed scene more closely on revisit, scoring 79.27 on GMS compared with 39.41 for Matrix-Game 3.0.
Their order reverses on camera rotation response, where they score 23.64 and 95.55, respectively (Table~\ref{tab:main_results}).
Matrix-Game 3.0 also scores below 10 on each of the physical causality, content causality, and evolution checks.
DreamX-World likewise follows rotation commands accurately (CR of 90.32) and outperforms Matrix-Game 3.0 on all three checks.
\subsection{Further Analysis}
\addtocontents{toc}{\protect\MainContentsDetail{4}}

\paragraph{Agreement between automatic and human evaluation.}
To assess whether automatic evaluation reproduces human model rankings, we sampled 40 tasks stratified by suite and target duration. Three annotators independently evaluated outputs from ten open-weight models.
For each metric, we aggregated automatic and human scores over the same evaluated videos within each model and compared the rankings of the ten models.
Spearman rank correlation was at least 0.900 for all twelve metrics and averaged 0.945 (Table~\ref{tab:human_model_rankings}). After mapping automatic scores to the human rating scale, the mean Pearson correlation between automatic and human model scores was 0.933.

\begin{table}[H]
\centering
\caption{
Model-level correlations between automatic and human evaluation. For each metric, we report Spearman rank correlation (SRCC) between the rankings of ten models and Pearson linear correlation (PLCC) between the mapped automatic scores and human scores for those models. The final column averages the twelve metric-specific correlations.
}
\label{tab:human_model_rankings}
\resizebox{\linewidth}{!}{%
\begin{tabular}{lccccccccccccc}
\toprule
Statistic
& PC & CC & EC & BC & VQ & GMS & LMS
& SC & TR & DD & CR & JP & Mean \\
\midrule
SRCC
& 0.976 & 0.952 & 0.952 & 0.952 & 0.945 & 0.939
& 0.915 & 0.921 & 0.929 & 0.900 & 0.975 & 0.982 & 0.945 \\
PLCC 
& 0.959 & 0.970 & 0.972 & 0.963 & 0.972 & 0.910
& 0.904 & 0.874 & 0.861 & 0.988 & 0.879 & 0.945 & 0.933 \\
\bottomrule
\end{tabular}}

\end{table}

\begin{figure}[H]
    \centering
    \includegraphics[width=\linewidth]
    {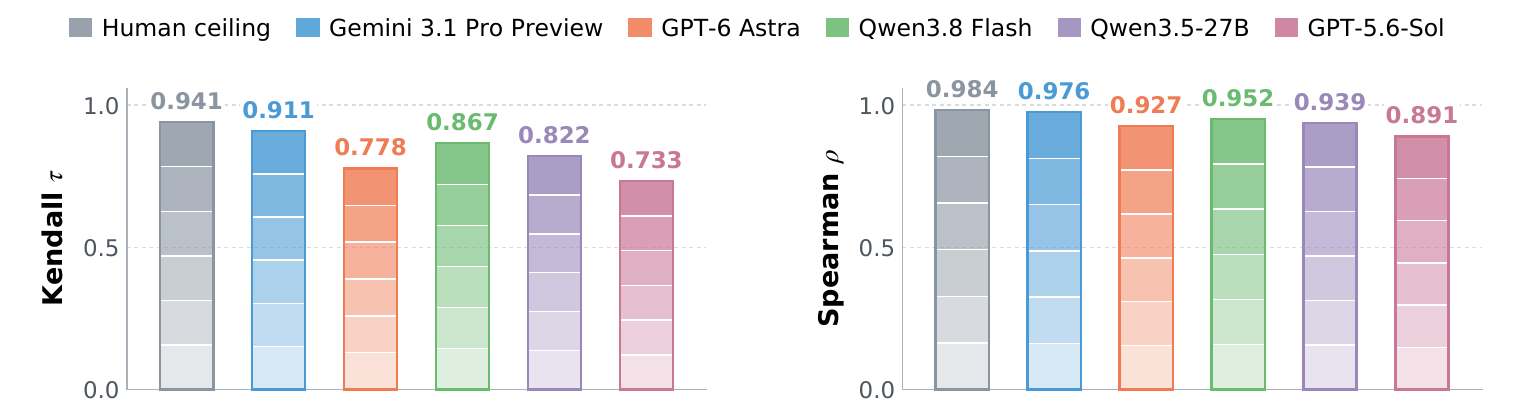}
    \caption{
    Agreement between five checklist judges and human annotators on model rankings by the PC/CC/EC composite score. The left panel shows Kendall's $\tau$ and the right panel shows Spearman's $\rho$. The human reference is the mean ranking correlation across the three annotator pairs.
    }
    \label{fig:judge_ranking}
\end{figure}

\paragraph{Agreement between checklist judges and human rankings.}
On the same tasks, we examined how agreement with the human ranking varied across checklist judges.
For each judge, we ranked models by the equally weighted mean of their physical causality (PC), content causality (CC), and evolution consistency (EC) scores and compared this ranking with the one derived from the three annotators' mean scores.
Of the five judges, the default judge, Gemini 3.1 Pro Preview, agreed most closely with the human ranking, with Kendall's $\tau=0.911$ and Spearman's $\rho=0.976$ (Figure~\ref{fig:judge_ranking}). For reference, mean pairwise agreement among the three annotators was 0.941 for $\tau$ and 0.984 for $\rho$.

\begin{figure}[H]
    \centering
    \begin{minipage}[t]{0.39\linewidth}
        \vspace{0pt}
        \centering
        \includegraphics[width=\linewidth]
            {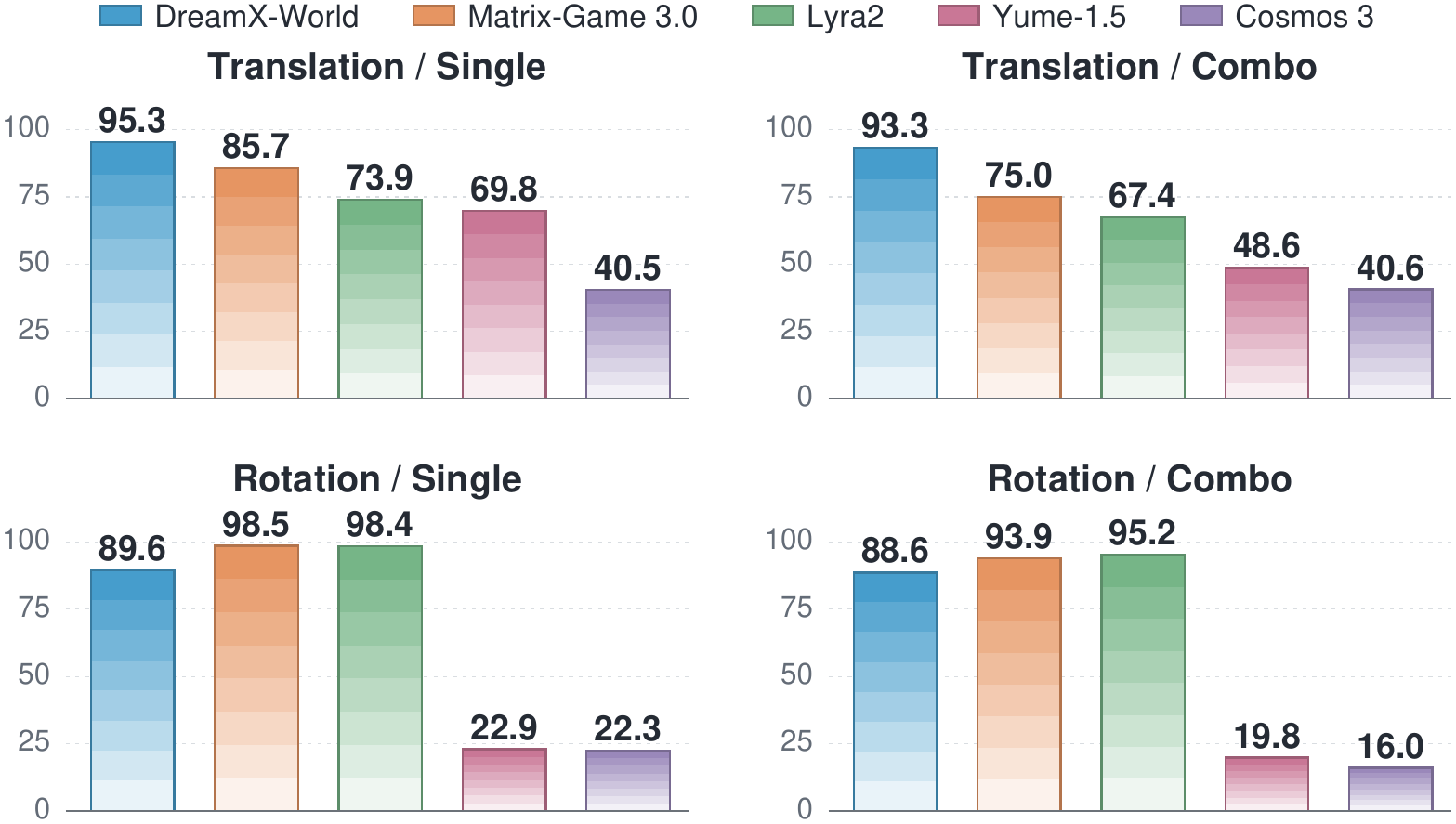}
        \caption{\textbf{Action-type analysis.}
        Control accuracy (\%) for translation (top) and rotation
        (bottom), under single-key (left) and composite-key
        (right) inputs.}
        \label{fig:actiontype}
    \end{minipage}\hfill%
    \begin{minipage}[t]{0.59\linewidth}
        \vspace{0pt}
        \centering
        \includegraphics[width=\linewidth]
            {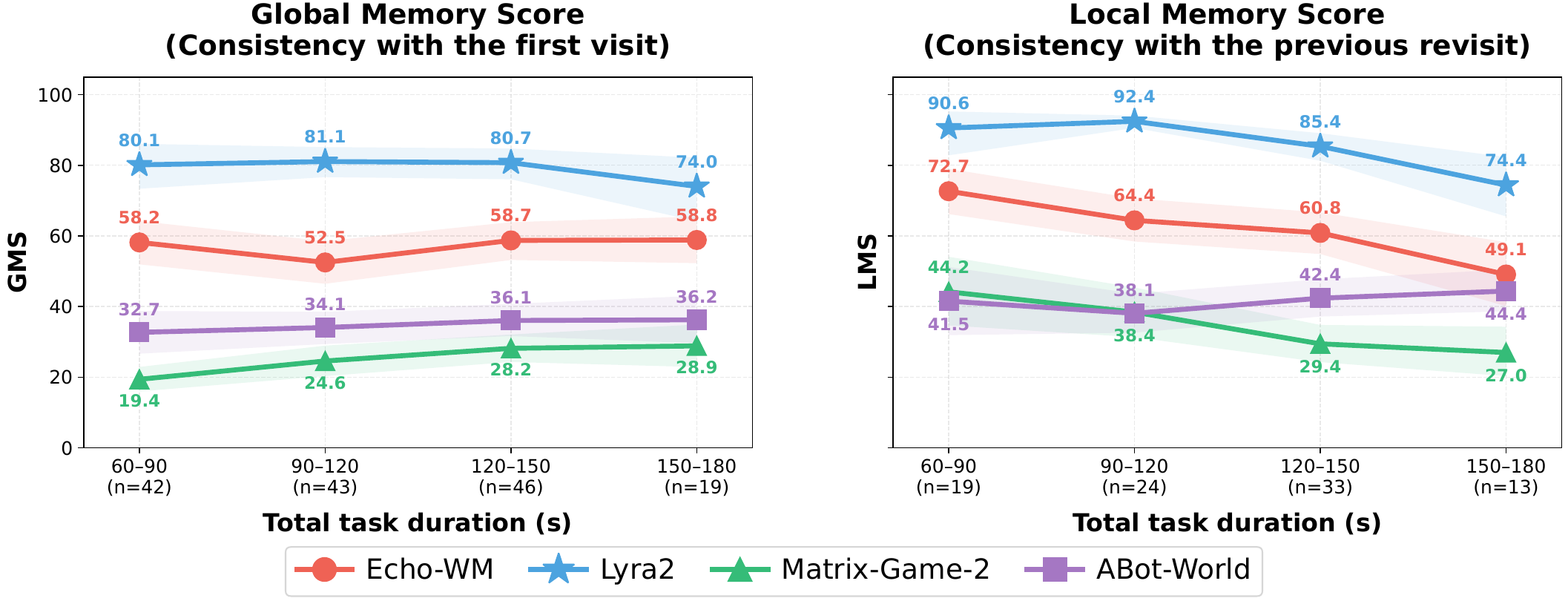}
            \vspace{-13pt}
        \caption{\textbf{Revisit consistency.}
        Mean GMS and LMS (0--100) across task durations.
        Shading shows 95\% bootstrap confidence intervals
        (10,000 case-level resamples); $n$ denotes the number
        of shared evaluation cases per group.}
        \label{fig:further_analysis}
    \end{minipage}
\end{figure}

\paragraph{Action-Type Analysis.}
Figure~\ref{fig:actiontype} shows that aggregate control scores can obscure differences between translation and rotation. Lyra 2.0 performs better on rotation, whereas Yume-1.5 performs better on translation. Combined inputs reduce translation accuracy for most of the selected models, while DreamX-World maintains strong performance across all four settings. These results distinguish accuracy on individual control types from the ability to execute combined commands. Appendix~\ref{app:translation_tolerance} compares single and composite inputs at a stricter translation tolerance. Appendix~\ref{app:stage_control} reports control performance across trajectory stages.

\paragraph{Long-Horizon Performance.}
Figure~\ref{fig:further_analysis} groups tasks by total duration and compares each revisit with the first observation (GMS) and with the preceding revisit (LMS).
The groups contain different scenes and trajectories. Three of the four models have lower LMS in the 150--180-second group than in the 60--90-second group.
Echo-WM shows the largest gap, scoring 72.7 in the shortest group and 49.1 in the longest, with GMS close to 58 in both.
For Matrix-Game 2.0 the two metrics move in opposite directions across these groups. Its LMS is lower in the longest group, while its similarity to the first observation is higher.
\vspace{10pt}

\paragraph{Metric Complementarity and Ranking Robustness.}
To examine how much information the metrics share, we computed pairwise Spearman correlations (Appendix~\ref{app:metriccorr}, Figure~\ref{fig:metriccorr}).
The mean absolute correlation was 0.22. The two revisit metrics, GMS and LMS, showed the strongest association ($\rho=0.76$).
Motion magnitude (DD) correlated negatively with subject consistency and both revisit metrics. In the evaluated samples, greater visual motion tended to accompany lower appearance or revisit similarity.
We also examined how control thresholds affected model rankings. Across 20 combinations of translation and rotation thresholds, correlations with the default ranking were at least 0.964, and the largest change was four places.
Appendix~\ref{app:composition_sensitivity} examines scene composition and category weights.

\paragraph{Audio Evaluation Results.}
We evaluate audio for Echo-WM and HappyOyster, which generate sound alongside video (Table~\ref{tab:audio_evaluation}).
Audio quality (AQ) and acoustic realism (AR) assess the soundtrack itself.
Using the corresponding video, we also measure semantic alignment between sound and scene (IB), predicted audio--video synchronization offset (SF), and whether visible actions have corresponding sound events (AAC).
Appendix~\ref{app:audio_evaluation} details the metrics and sampling procedures.

\par
\begingroup
\setlength{\columnsep}{10pt}
\setlength{\intextsep}{4pt}
\begin{wraptable}{r}{0.5\columnwidth}
\centering
\vspace{2pt}
\caption{Audio evaluation results for Echo-WM and HappyOyster.
Higher is better except for SF.}
\label{tab:audio_evaluation}
\vspace{-4pt}
\scriptsize
\renewcommand{\arraystretch}{1.08}
\resizebox{\linewidth}{!}{%
\begin{tabular}{@{}lccccc@{}}
\toprule
Model
& AQ$\uparrow$
& IB$\uparrow$
& SF$\downarrow$
& AR$\uparrow$
& AAC$\uparrow$ \\
\midrule
Echo-WM     & 6.573 & 0.1545 & 1.039 & 44.09 & 35.22 \\
HappyOyster & 5.641 & 0.1354 & 0.912 & 48.09 & 28.91 \\
\bottomrule
\end{tabular}%
}
\end{wraptable}

\noindent
Echo-WM achieves higher audio quality, audio--visual semantic alignment,
and action--sound correspondence, whereas HappyOyster obtains higher
acoustic realism and a lower predicted synchronization offset.
Later portions of some rollouts contain noise, temporal drift, and repeated
synthetic sound patterns. Footsteps often follow the walking motion approximately,
with less consistent timing for other sound events.
\par
\endgroup

\subsection{Discussion of Error Patterns}
\addtocontents{toc}{\protect\MainContentsDetail{2}}

We use Gemini 3.1 Pro Preview to annotate 500 videos generated by
five models on the same 100 tasks using 25 predefined error labels.
Each video is represented by 20 uniformly sampled frames, each tagged
with its source timestamp, together with the camera-control
instructions. The judge outputs video-level failure labels only and
does not produce temporal localizations. The complete prompt is
provided in Appendix~\ref{app:failure_type_prompt}.

\begin{figure}[H]
    \centering
    \includegraphics[width=\linewidth]
    {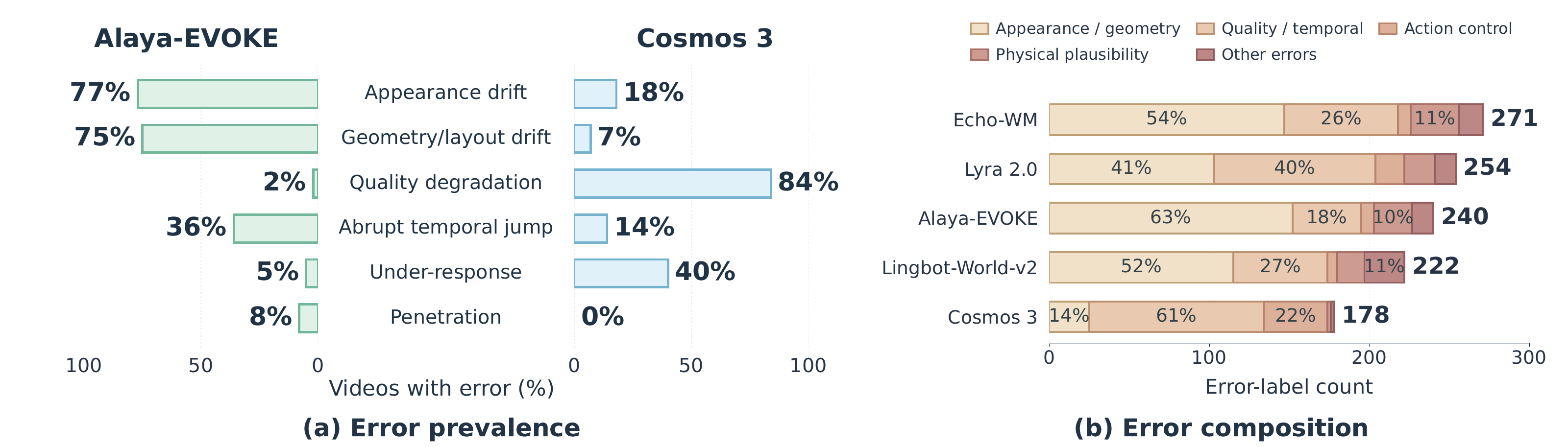}
    \caption{Error analysis on five models.
(a) Video-level prevalence of the six most frequent pooled labels
for two contrasting models.
(b) Error-label counts and category composition across all five
models; internal percentages denote shares of each model's total
labels. See more in Figure~\ref{fig:error-appendix}.}
\label{fig:error-main}
\end{figure}

\paragraph{Aggregate counts obscure distinct failure profiles.}
On the same 100 tasks, Cosmos~3 averaged 1.78 error labels per video, compared with 2.40 for Alaya-EVOKE.
The distribution by error type reveals different weaknesses (Figure~\ref{fig:error-main}).
Quality degradation occurred in 84\% of Cosmos~3 videos and 2\% of Alaya-EVOKE videos.
Errors for Alaya-EVOKE concentrated on scene retention, with appearance drift in 77\% of videos and geometry or layout drift in 75\%.

\paragraph{Appearance and geometry failures frequently coincide.}
Within individual videos, appearance drift frequently accompanied geometry or layout drift.
Across the 500 videos generated by five models, 203 carried both labels, accounting for 60\% of the videos with at least one of the two errors (Jaccard coefficient of 0.60, Figure~\ref{fig:error-appendix}(e)).
Several error types can also be frequent within a model's outputs. Lyra~2.0, for example, frequently exhibited both scene drift and quality degradation across its videos.

\section{Conclusion}

We introduced LongWM-Bench to evaluate whether interactive world models remain controllable and preserve previously observed scenes during extended interaction. The benchmark comprises 300 tasks with target durations of 60–180 seconds, covering sustained control and delayed and repeated revisits. Our evaluation reveals that similar appearance consistency can mask substantial differences in revisit retention, while accurate camera responses can coexist with weak physical and causal validity. Coherent scene evolution remains a common weakness across the evaluated models. These findings highlight the need to assess control, persistence, and evolution together when evaluating interactive worlds. LongWM-Bench provides a diagnostic framework for this purpose and supports progress toward generated environments that remain reliable throughout continued interaction.

\bibliographystyle{assets/plainnat}
\bibliography{references}
\clearpage
\addtocontents{toc}{\protect\AppendixContentsStart}
\section*{Appendix Contents}
\begingroup
  \setcounter{tocdepth}{-5}
  \def\AppendixContentsStart{\setcounter{tocdepth}{3}}
  \sffamily\fontsize{9.8}{10.7}\selectfont
  \color{LinkPurple}\hypersetup{linkcolor=LinkPurple,linktoc=all}
  \makeatletter
  \@starttoc{toc}
  \makeatother
  \setcounter{tocdepth}{2}
\endgroup
\clearpage
\appendix




\section{Human Annotation Protocol}

\label{sec:human_annotation_protocol}

\begin{figure*}[htbp]
    \centering
    \includegraphics[width=\textwidth]{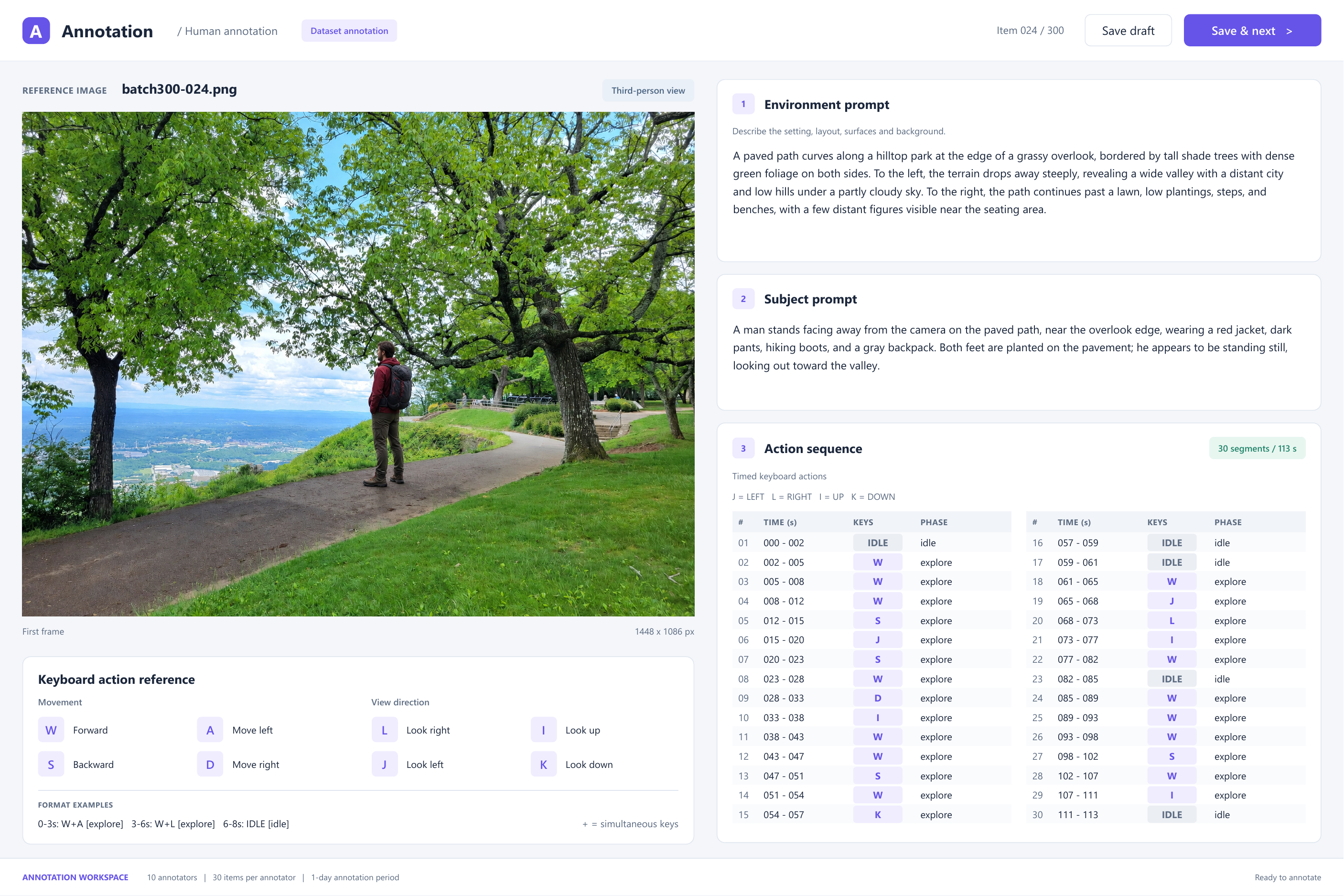}
    \caption{Human annotation interface for the 300-item annotation
    task. The interface displays the current reference image,
    environment and subject prompts, and a time-indexed action
    sequence, together with the WASD movement and LIJK view-control
    mappings.}
    \label{fig:annotation_interface}
\end{figure*}

We recruited 10 human annotators to annotate a total of 300 items,
with 30 items assigned to each annotator. The annotation task was
completed within one day, and each annotator received US\$50 for
their work. As shown in Figure~\ref{fig:annotation_interface},
the annotation interface presents one reference image at a time,
together with fields for an environment prompt, a subject prompt,
and a time-indexed action sequence. The environment prompt
describes the scene layout and background, while the subject
prompt specifies the controlled subject's appearance, position,
and relationship to the scene. The action sequence specifies
the start and end times of each segment and its associated
keyboard inputs: \texttt{W}, \texttt{A}, \texttt{S}, and \texttt{D}
encode movement, while \texttt{L}, \texttt{I}, \texttt{J}, and
\texttt{K} encode rightward, upward, leftward, and downward view
changes, respectively. Simultaneous inputs are joined by
\texttt{+}, and \texttt{IDLE} denotes a segment without input.
Annotators save the current annotation before advancing to
the next assigned item.

\section{Revisit-Memory Suite Analysis}
\label{app:revisit_suite}

The Revisit-Memory suite contains 150 predefined trajectories.
To characterize different forms of spatial return and memory demand,
we group these trajectories into six path types. We also compare model rankings
under pose-based return criteria.

\subsection{Six Revisit Path Types}
\label{app:revisit_path_types}

We categorize revisit paths into the following six types, covering both simple viewpoint changes and complex multi-waypoint navigation.

\begin{itemize}
    \item \textbf{In-place Turn Revisit (ITR):} The camera rotates in place to revisit surrounding views, including full $360^\circ$ panning.
    Typical patterns: $L \rightarrow R$, $L \rightarrow R \rightarrow L$, or a full $360^\circ$ yaw rotation.

    \item \textbf{Step--Look--Return (SLR):} The agent steps forward, looks around, and then steps back to the original position.
    Typical pattern: $W \rightarrow$ look left/right $\rightarrow S$.

    \item \textbf{Radial Revisit (RR):} The agent departs from a central point in multiple directions and returns to the same center.
    Typical patterns: $W \rightarrow S$, $A \rightarrow D$, $W+A \rightarrow S+D$, all starting from the same origin.

    \item \textbf{Same-path Round-trip Revisit (SRR):} The agent follows a path and returns along the exact same route.
    Typical patterns: $W \rightarrow S$, $A \rightarrow D$, $L \rightarrow R$, or longer symmetric sequences.

    \item \textbf{Different-path Loop Revisit (DLR):} The agent returns to the starting area via a different route, forming a closed loop.
    Typical pattern: $W \rightarrow A \rightarrow S \rightarrow D$, or other non-reversing loops.

    \item \textbf{Multi-waypoint Interleaved Revisit (MIR):} The agent visits multiple waypoints in sequence and later revisits them in an interleaved order.
    Typical patterns: $A \rightarrow B \rightarrow C \rightarrow B \rightarrow A$, or $A \rightarrow B \rightarrow A \rightarrow B$.

\end{itemize}

\begin{figure*}[t]
    \centering
    \includegraphics[width=\textwidth]{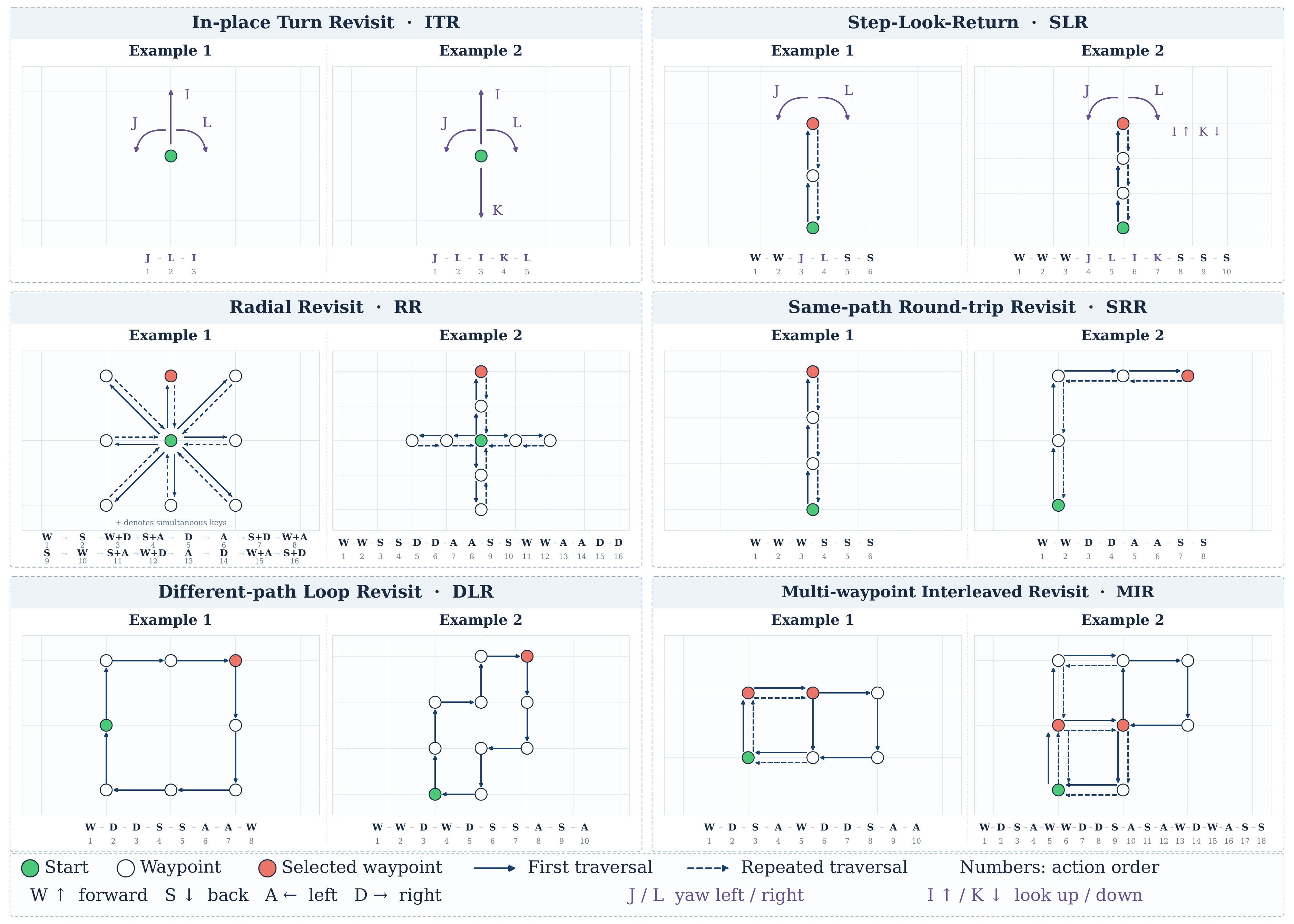}
    \caption{
        Six categories of revisit trajectories, with two examples per category.
        WASD controls translation, while IJKL controls camera rotation.
        Each action sequence corresponds to the illustrated trajectory.
        Green, white, and red circles denote starting positions,
        intermediate waypoints, and selected waypoints, respectively.
        Solid and dashed arrows indicate first and repeated traversals
        of a path segment; curved arrows indicate camera rotation.
    }
    \label{fig:revisit_trajectories}
\end{figure*}

\subsection{Pose-Conditioned Revisit Evaluation}
\label{app:pose_conditioning}

The original GMS/LMS scores jointly reflect navigation
execution and visual retention at scheduled revisit observations. A successful revisit requires both
leaving and returning to an anchor and preserving its scene content;
visual similarity alone does not identify which capability failed.

To assess sensitivity to return-pose differences, we compute
$\mathrm{GMS}_{\mathrm{cond}}$ and
$\mathrm{LMS}_{\mathrm{cond}}$ using ViPE-based pose verification.
A case is retained only when every revisit observation satisfies
the rotation and translation tolerances relative to its corresponding
initial anchor observation. Cases failing these conditions or
lacking usable poses are excluded from the conditional calculation.
Observation frames, visual features, and aggregation weights remain
unchanged. This case-level filter assesses return-pose agreement;
the shared-pair audit below additionally checks sustained departure
and approximate return on identical effective samples.

Table~\ref{tab:pose_conditioning} compares the conditional and
original rankings across twelve tolerance settings. Correlations
are computed over models with both scores available at each setting.
Spearman correlations range from 0.771 to 1.000 for GMS and from
0.900 to 1.000 for LMS, with a maximum GMS rank displacement of
two positions. The broad model comparisons remain consistent under
the tested filters. The conditional scores characterize visual
retention on pose-qualified revisits, with retained case sets
potentially differing across models.

\begin{table}[t]
\centering
\caption{Sensitivity of model rankings to return-pose conditioning.
Translation tolerances are expressed in ViPE coordinate units.
Rank-displacement and coverage columns refer to GMS.}
\label{tab:pose_conditioning}
\resizebox{\linewidth}{!}{%
\begin{tabular}{cccccccccc}
\toprule
Rotation & Translation
& $\rho$ (GMS) & $\tau_b$ (GMS)
& $\rho$ (LMS) & $\tau_b$ (LMS)
& \makecell{Max shift\\(GMS)}
& \makecell{Mean shift\\(GMS)}
& \makecell{Models\\(GMS)}
& \makecell{Mean cases/model\\(GMS)} \\
\midrule
45$^\circ$ & 1 & 0.900 & 0.800 & 0.900 & 0.800 & 1 & 0.4 & 5 & 27 \\
45$^\circ$ & 2 & 1.000 & 1.000 & 1.000 & 1.000 & 0 & 0.0 & 6 & 34 \\
45$^\circ$ & 5 & 1.000 & 1.000 & 1.000 & 1.000 & 0 & 0.0 & 6 & 54 \\
40$^\circ$ & 1 & 0.900 & 0.800 & 0.900 & 0.800 & 1 & 0.4 & 5 & 27 \\
40$^\circ$ & 2 & 0.771 & 0.600 & 1.000 & 1.000 & 2 & 0.7 & 6 & 33 \\
40$^\circ$ & 5 & 1.000 & 1.000 & 0.943 & 0.867 & 0 & 0.0 & 6 & 52 \\
35$^\circ$ & 2 & 0.943 & 0.867 & 1.000 & 1.000 & 1 & 0.3 & 6 & 31 \\
35$^\circ$ & 5 & 1.000 & 1.000 & 0.943 & 0.867 & 0 & 0.0 & 6 & 50 \\
30$^\circ$ & 2 & 0.900 & 0.800 & 1.000 & 1.000 & 1 & 0.4 & 5 & 36 \\
30$^\circ$ & 5 & 0.943 & 0.867 & 0.943 & 0.867 & 1 & 0.3 & 6 & 47 \\
25$^\circ$ & 2 & 0.900 & 0.800 & 1.000 & 1.000 & 1 & 0.4 & 5 & 33 \\
25$^\circ$ & 5 & 1.000 & 1.000 & 0.900 & 0.800 & 0 & 0.0 & 5 & 52 \\
\bottomrule
\end{tabular}}
\end{table}

\paragraph{Departure-and-return audit on shared observation pairs.}

We audit Lyra 2.0, Echo-WM, LingBot-World-v2, and Matrix-Game 2.0
using ViPE trajectories from 150 revisit tasks.
Departure requires a rotation change exceeding $30^\circ$ or
a translation change exceeding 2 ViPE units, sustained for at least
0.5 seconds and three consecutive valid pose samples.
Both observations in a pair must also lie within the specified
return tolerances of the same anchor's initial pose.
We retain only observation pairs that pass both checks for all four
models and recompute GMS$'$ and LMS$'$ with the original scoring
and aggregation rules.
At return tolerances of $60^\circ$/15 units, we retain 82 GMS pairs
from 51 tasks and 25 LMS pairs from 21 tasks.
Both metrics preserve the four-model rankings
(Spearman $\rho=1.000$) relative to both the full common baseline
and the baseline restricted to retained tasks, each computed
before pair filtering.
The same ranking agreement holds at $45^\circ$/10 units
and $60^\circ$/10 units.

\section{Further Analysis}
\subsection{Metric Redundancy and Score Robustness}
\label{app:metriccorr}
Pairwise Spearman correlations indicate limited redundancy among the metrics (Figure~\ref{fig:metriccorr} in Appendix~\ref{app:metriccorr}), with a mean absolute correlation of only $0.22$ across valid off-diagonal pairs. The strongest correlation is between GMS and LMS ($\rho=0.76$), as expected from their shared DINOv2-based similarity formulation, followed by PC--CC ($0.67$) and TR--CR ($0.60$), both within semantically related dimensions. Overall, within-dimension correlations are higher than cross-dimension correlations ($0.38$ vs.\ $0.20$), suggesting that the five dimensions are internally coherent yet largely complementary. Dynamic Degree is a notable exception, showing negative correlations with several consistency and memory metrics (e.g., SC $-0.48$, JP $-0.34$, GMS/LMS $-0.30/-0.34$), highlighting a motion--consistency trade-off and motivating joint reporting of these metrics. The evaluation is also robust to metric parameterization: sweeping the translation tolerance ($40^\circ$--$80^\circ$) and rotation threshold ($2^\circ$--$8^\circ$) over 20 settings preserves model rankings with $\rho \ge 0.964$ and shifts no model by more than four positions.
\begin{figure}[t]
  \centering
  \includegraphics[width=\linewidth]{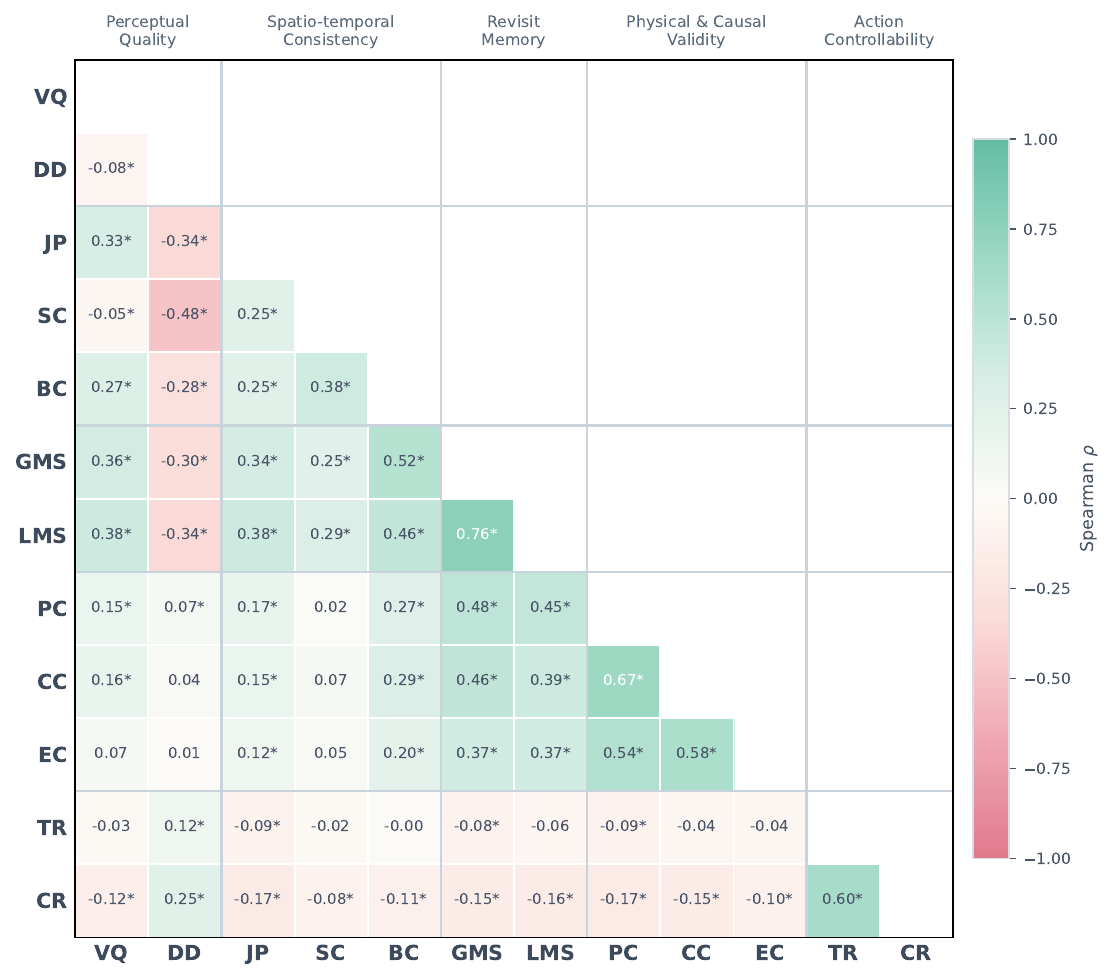}
  \caption{\textbf{Metric correlation.}}
  \label{fig:metriccorr}
\end{figure}

\subsection{Additional Error Analysis}
\begin{figure}[t]
    \centering
    \vspace{-20pt}
    \includegraphics[width=\linewidth]
    {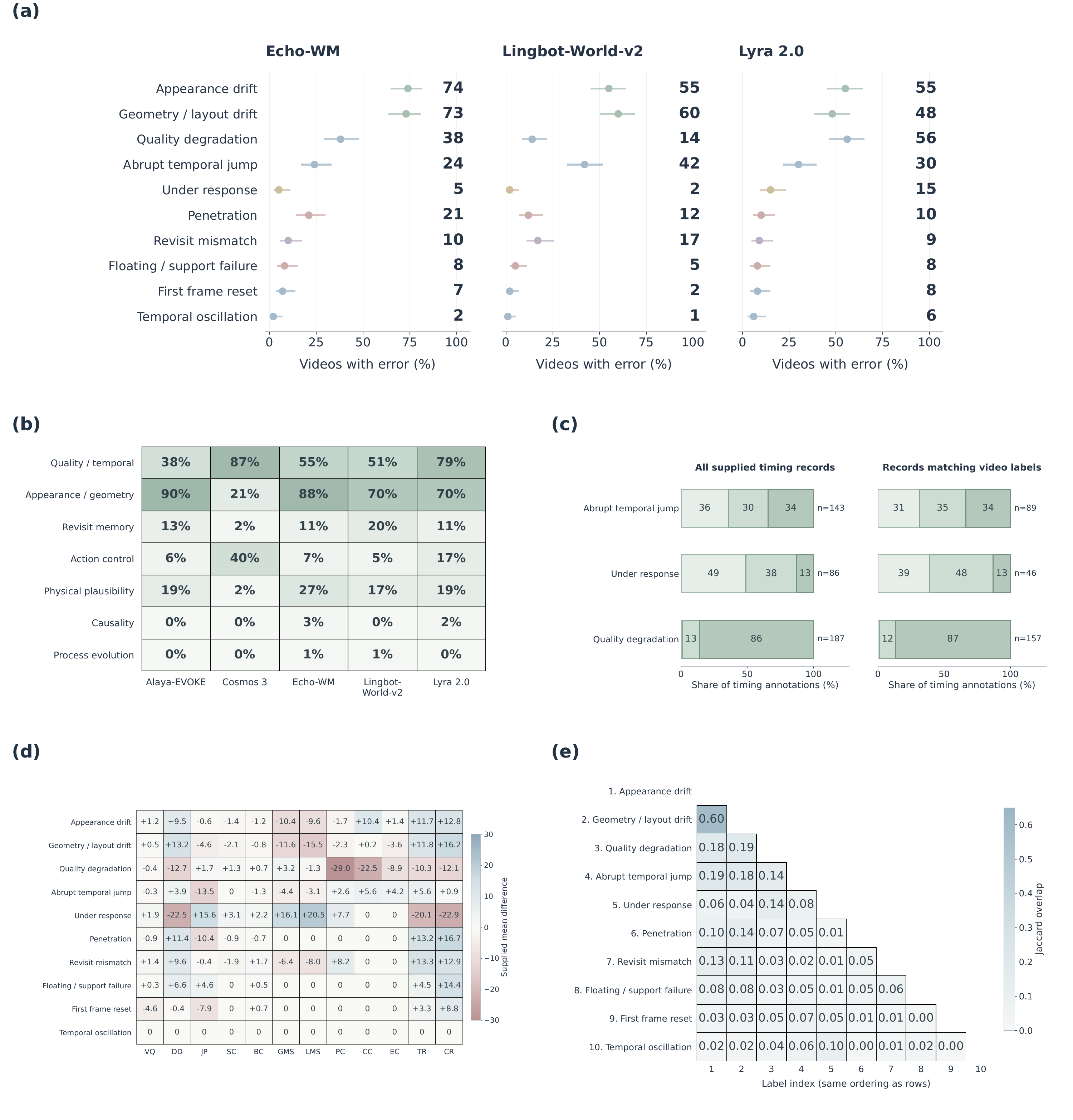}
    \vspace{-10pt}
    \caption{Supplementary error analysis.
(a) Remaining model profiles with marginal 95\% Wilson intervals.
(b) Video-level category prevalence, deduplicated within categories.
(c) Early/middle/late distributions from a separate auxiliary
temporal-localization annotation, shown for all timing records
(left) and records matching the video-level labels (right);
$n$ denotes the annotation count.
(d) Mean metric differences between videos with and without each label.
(e) Pairwise Jaccard overlap of the ten most frequent labels.}
\label{fig:error-appendix}
    \vspace{-5pt}
\end{figure}
Figure~\ref{fig:error-appendix} extends the main-text analysis
with the remaining model profiles, deduplicated category
prevalence, temporal annotations, metric associations, and label
co-occurrence. Echo-WM exhibits substantial appearance and
geometry/layout drift (74\% and 73\%), whereas LingBot-World-v2
has the highest observed prevalence of abrupt temporal jumps
(42\%). Lyra~2.0 exhibits a mixed profile, combining quality
degradation (56\%) with appearance and geometry/layout drift
(55\% and 48\%).

Panel~\ref{fig:error-appendix}(c) uses a separate annotation of early,
middle, and late events. After normalizing one label spelling, 772 of the
1,114 timing records matched the video-level labels. Quality degradation
occurred late in 86.1\% of all timing records and 86.6\% of matched records.
Under-response was most frequent in the early stage across all records and
in the middle stage in the matched subset. Videos labeled with under-response
had lower mean TR and CR. Videos with geometry or layout drift had lower
mean GMS and LMS.

\newpage
\subsection{Translation-Tolerance Sensitivity}
\label{app:translation_tolerance}

TR compares net camera displacement with the sum of the
commanded translation vectors. The default $60^\circ$ tolerance
accepts net displacement within the angular tolerance. Executing only
W for W+D gives a $45^\circ$ deviation and is counted as correct.
We therefore repeat the analysis at $30^\circ$, where this partial
execution fails, using eleven models excluding Genie 3.
Evaluable action segments are fixed within each model and input group.

\begin{figure}[!htbp]
\centering
\includegraphics[width=\linewidth]
{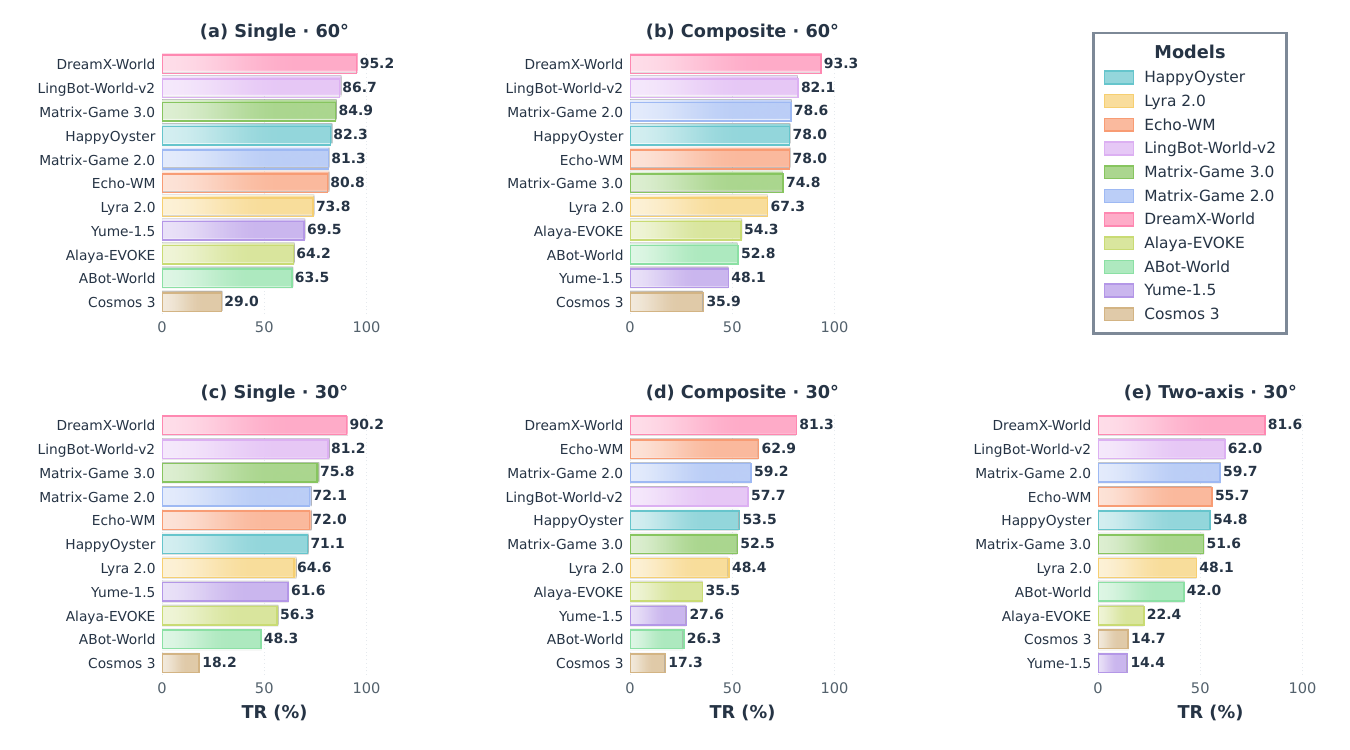}

\caption{
Translation accuracy under default and stricter directional
criteria (\%). Panels (a--b) use the default $60^\circ$ tolerance;
panels (c--e) use $30^\circ$.
Scores equally weight action segments within each input group.
Composite inputs include additional translation or rotation keys;
two-axis inputs contain two translation directions and may also
include rotation. Models are sorted independently within each
panel, with consistent colors across panels.
}
\label{fig:composite_strict}
\end{figure}

Figure~\ref{fig:composite_strict} shows that all eleven models
have lower TR on composite-key and two-axis translation inputs
than on single-key inputs under the stricter criterion.
At this tolerance, DreamX-World scored 90.2\% on single-key inputs,
81.3\% on composite inputs, and 81.6\% on two-axis inputs.

\begin{figure}[!htbp]
    \centering
    \includegraphics[width=0.8\linewidth]
    {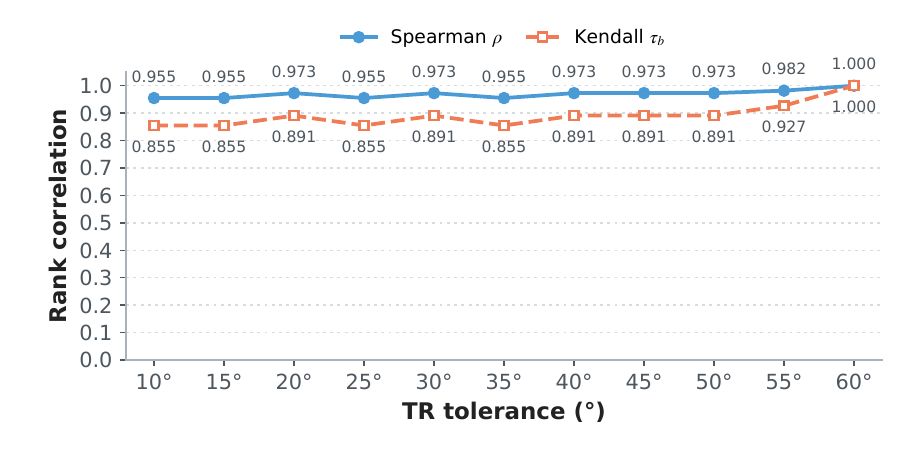}
    \caption{
    Model-ranking sensitivity to the TR directional tolerance
    across eleven models. Spearman's $\rho$ and Kendall's
    $\tau_b$ compare each ranking with the default $60^\circ$
    ranking. The maximum rank displacement across the tested
    thresholds is two positions.
    }
    \label{fig:tr_sweep}
\end{figure}

We additionally sweep the tolerance from $60^\circ$ to
$10^\circ$ in $5^\circ$ steps, keeping each model's evaluable
videos and action segments fixed. Overall TR uses equally
weighted task means. Figure~\ref{fig:tr_sweep} compares the
resulting model rankings with the default $60^\circ$ ranking.
The $30^\circ$ ranking has $\rho=0.973$; the minimum across
the sweep is 0.955, with a maximum rank displacement of
two positions. These results indicate stable model comparisons
under stricter directional criteria.

\FloatBarrier
\subsection{Stage-Wise Control Performance}
\label{app:stage_control}

\begin{table}[htbp]
\centering
\captionsetup{font=small,skip=4pt}
\caption{
Early--late control agreement across ten models.
}
\label{tab:early_late_control}

\small
\begin{tabular*}{\linewidth}
{@{\extracolsep{\fill}}lcc@{}}
\toprule
Metric & Tasks & Spearman $\rho$ \\
\midrule
TR & 110 & 0.648 \\
CR & 133 & 0.896 \\
\bottomrule
\end{tabular*}
\end{table}

We divide each trajectory into four equal temporal stages
and examine the relationship between early directional
responsiveness and late-stage instruction following.
Late-stage success requires all five final relevant instructions
in the last stage, including combined controls, to satisfy
the corresponding TR/CR criterion.
Across ten models, we correlate this success rate with
first-stage scores using matched tasks and action segments
and equal task weights.
Table~\ref{tab:early_late_control} shows positive correlations
of 0.648 for TR and 0.896 for CR, covering 110 of 158 eligible
tasks and 133 of 186, respectively.
The early and late measurements use non-overlapping stages.
These results link early directional responsiveness to
subsequent multi-instruction execution.

Across ten models, first- and final-stage rankings were positively correlated on tasks with relevant action segments in every stage. Using segments common to all ten models, Spearman correlation was 0.806 for translation response (TR) across 180 tasks and 0.891 for rotation response (CR) across 175 tasks. The change in accuracy varied by model. HappyOyster's TR fell from 80.5\% to 78.5\%, while Echo-WM's fell from 92.8\% to 75.3\%.

For Echo-WM, translation accuracy declined most on commands combining translation and rotation, from 91.2\% in the first stage to 64.6\% in the final stage. On translation-only commands, TR fell from 92.6\% to 75.9\%. Rotation responses were more stable. Yaw-only CR remained at 100\%, and pitch-only CR changed from 98.7\% to 96.9\%. The comparisons for translation alone, combined inputs, yaw alone, and pitch alone covered 125, 97, 52, and 50 matched tasks, respectively.

\subsection{Data-Composition Sensitivity}
\label{app:composition_sensitivity}

We stratify cases by scene source
(real, real-AI-edited, and AI-generated), viewpoint
(first- and third-person), and scene domain.
Within each subset, model scores are recomputed and their
rankings are compared with the corresponding full-data rankings.
Table~\ref{tab:composition_subsets} reports minimum and median Spearman
correlations across subsets within each stratification.
GMS/LMS rankings remain particularly stable, with minimum
correlations of 0.94--0.99 and median correlations of 0.98--1.00.
Greater variation is observed for TR across viewpoint subsets
(minimum/median $\rho = 0.83/0.87$) and for JP and EC across
scene domains ($0.59/0.88$ and $0.73/0.90$, respectively).

\begin{table}[t]
\centering
\caption{
Model-ranking stability across benchmark subsets.
For each stratification, the upper and lower rows report
the median and minimum Spearman correlations with
the corresponding full-data model rankings, respectively.
}
\label{tab:composition_subsets}
\begingroup
\footnotesize
\setlength{\tabcolsep}{2pt}
\renewcommand{\arraystretch}{1.08}

\begin{tabular*}{\linewidth}
{@{\extracolsep{\fill}}l*{12}{c}@{}}
\toprule
Stratification
& VQ & DD & JP & SC & BC & GMS & LMS
& PC & CC & EC & TR & CR \\
\midrule
\multirow{2}{*}{Scene source}
& 0.98 & 0.98 & 0.96 & 0.95 & 0.98 & 0.99 & 1.00
& 0.95 & 0.97 & 0.94 & 0.98 & 0.98 \\
& 0.97 & 0.97 & 0.93 & 0.93 & 0.97 & 0.99 & 0.99
& 0.92 & 0.97 & 0.90 & 0.97 & 0.97 \\
\midrule
\multirow{2}{*}{Viewpoint}
& 0.98 & 0.98 & 0.96 & 0.93 & 0.95 & 0.98 & 1.00
& 0.97 & 0.96 & 0.97 & 0.87 & 0.96 \\
& 0.96 & 0.97 & 0.95 & 0.91 & 0.92 & 0.97 & 0.99
& 0.97 & 0.95 & 0.96 & 0.83 & 0.92 \\
\midrule
\multirow{2}{*}{Domain}
& 0.98 & 0.97 & 0.88 & 0.94 & 0.95 & 0.98 & 0.99
& 0.95 & 0.94 & 0.90 & 0.95 & 0.97 \\
& 0.90 & 0.94 & 0.59 & 0.92 & 0.89 & 0.94 & 0.95
& 0.90 & 0.90 & 0.73 & 0.94 & 0.90 \\
\bottomrule
\end{tabular*}

\endgroup
\end{table}

To assess the effect of category proportions, we additionally
assign equal total weight to each source category, viewpoint,
or domain. Each comparison retains the eligible case population
and model set used by its original-weight reference.
Table~\ref{tab:composition_weights} reports correlations of
0.97--1.00, supporting stable model comparisons under the
tested reweightings.

\begin{table}[t]
\centering
\caption{
Ranking correlations under uniform category weighting
versus original weighting.
}
\label{tab:composition_weights}
\resizebox{\linewidth}{!}{%
\begin{tabular}{lrrrrrrrrrrrr}
\toprule
Weighting
& VQ & DD & JP & SC & BC & GMS & LMS
& PC & CC & EC & TR & CR \\
\midrule
Uniform scene source
& 1.00 & 1.00 & 1.00 & 1.00 & 1.00 & 1.00 & 1.00
& 1.00 & 1.00 & 1.00 & 1.00 & 0.99 \\
Uniform viewpoint
& 0.99 & 0.99 & 0.99 & 0.99 & 1.00 & 1.00 & 0.99
& 1.00 & 1.00 & 1.00 & 1.00 & 1.00 \\
Uniform domain
& 1.00 & 0.99 & 0.97 & 0.99 & 1.00 & 0.99 & 0.99
& 1.00 & 1.00 & 0.97 & 0.99 & 1.00 \\
\bottomrule
\end{tabular}}
\end{table}

\section{Inference Protocols and Model Configurations}
\label{app:video-inference}

The evaluated models differ in their conditioning inputs and control
interfaces. We therefore convert each benchmark case into the native
inference format of each model, preserving the reference image, textual
context, prescribed controls, and requested duration whenever supported.
Table~\ref{tab:model_inference_configs} summarizes the configurations
used in our evaluation.

For models with native keyboard, mouse, or action interfaces, the
benchmark controls are mapped directly to the corresponding movement
and camera commands. For trajectory-conditioned models, the prescribed
controls are converted into their native camera representations,
including camera poses, Pl{\"u}cker embeddings, PRoPE, or UCPE where
applicable. Translation and rotation are applied jointly when
simultaneous controls are supported, while intervals without prescribed
commands are represented by neutral or zero-motion inputs.

Several model-specific adaptations are necessary because the available
interfaces are not identical. Genie 3 is evaluated through the hosted
Project Genie interface, where controls are issued through browser
automation. The deployed interface does not expose direct control over
the underlying checkpoint, random seed, sampling steps, or guidance
scale. Matrix-Game 2.0 is evaluated with image conditioning and
keyboard/mouse actions without additional textual conditioning, whereas
Matrix-Game 3.0 additionally accepts a text prompt. DreamX-World
represents the prescribed actions through camera motion and therefore
does not provide an independently commanded character state. Lyra 2.0
similarly uses camera trajectories; for third-person cases, it
constructs a look-at trajectory around a proxy subject position.
Cosmos~3 is evaluated through its forward-dynamics action interface:
yaw is supported, whereas pitch has no corresponding action in this
configuration, and turning in place is approximated with a small
forward motion.

Natural-language interventions also differ across interfaces.
LingBot-World-v2 and Echo-WM can incorporate event descriptions as
global text conditioning, but the descriptions are not activated at
their prescribed execution times. We therefore denote these interfaces
as \textit{Global}. For models whose evaluated interfaces do not support
timed natural-language interventions, such cases are treated as
unsupported rather than approximated using unrelated movement or camera
controls.

\begin{table*}[tb!]
\centering
\caption{
Inference configurations of the evaluated interactive world models.
``Conditioning'' denotes the input reference image (Img.) and text,
while ``Control'' denotes the native interface used for benchmark
actions. ``--'' indicates unavailable or unspecified information.
Latency is the average inference time per case; resolution and FPS
refer to generated videos used for evaluation.
}
\label{tab:model_inference_configs}

\vspace{-2pt}

\begingroup
\fontsize{8.2pt}{9.2pt}\selectfont
\renewcommand{\arraystretch}{1.1}
\setlength{\tabcolsep}{2.5pt}

\begin{tabular*}{\textwidth}{
@{\extracolsep{\fill}}
lccccc
@{}
}
\toprule
\textbf{Model}
& \textbf{Conditioning}
& \textbf{Control}
& \textbf{Latency (s)}
& \textbf{Resolution}
& \textbf{FPS} \\
\midrule

Genie 3~\citep{genie3}
& Img. + Text & Keyboard & \LWlock
& $\sim1280\times704$ & 16 \\

HappyOyster~\citep{happyoyster}
& Img. + Text & Action & \LWlock
& $\sim832\times480$ & 24 \\

LingBot-World-v2~\citep{lingbotworldv2}
& Img. + Text & Camera pose & 587
& $\sim832\times480$ & 16 \\

Matrix-Game 2.0~\citep{he2025matrix}
& Img. & Keyboard + Mouse & 160
& $640\times352$ & 12 \\

Matrix-Game 3.0~\citep{wang2026matrixgame30realtimestreaming}
& Img. + Text & Keyboard + Mouse & 550
& $1280\times704$ & 16 \\

Yume-1.5~\citep{mao2026yume1}
& Img. + Text & Control tokens & 1802
& $1280\times704$ & 16 \\

DreamX-World~\citep{team2026dreamx}
& Img. + Text & Camera pose & 769
& $1280\times704$ & 16 \\

ABot-World~\citep{jiang2026abot}
& Img. + Text & Keyboard & 191
& $832\times480$ & 12 \\

Cosmos~3~\citep{nvidia2026cosmos3}
& Img. + Text & Relative pose & 1480
& 480p short side & 16 \\

Echo-WM~\citep{zhang2026echowmopenenterableomnimodal}
& Img. + Text & Camera trajectory & 1134
& $1280\times704$ & 24 \\

Alaya-EVOKE~\citep{yin2026alaya}
& Img. + Text & Camera pose & 962
& $640\times384$ & 24 \\

Lyra 2.0~\citep{shen2026lyra}
& Img. + Text & Camera pose & 6446
& $832\times480$ & 16 \\

\bottomrule
\end{tabular*}
\endgroup

\vspace{-4pt}
\end{table*}

\paragraph{Control alignment.}\label{app:control_alignment}

Each task specifies a shared navigation sequence comprising
translation, yaw/pitch rotation, combined inputs, idle intervals,
and segment durations.
Model-specific adapters convert this sequence into native key
states, mouse increments, action codes, or camera trajectories,
using fixed control magnitudes where configurable.
Segment durations are converted into control frames or action
blocks according to each interface's frame rate and generation
block length.
Table~\ref{tab:control-adapter-configs} summarizes the control
parameters, timing conversions, and interface adaptations,
showing how the shared navigation instructions and their intended
timing are implemented for each model.
Where supported, combined inputs are encoded jointly.
Idle intervals use the corresponding neutral inputs, with motion
during these intervals governed by the model's native dynamics.

\begin{table*}[t]
\centering
\footnotesize
\setlength{\tabcolsep}{4pt}
\begin{tabularx}{\linewidth}{@{}p{0.19\linewidth}p{0.29\linewidth}X@{}}
\toprule
\textbf{Model} &
\textbf{Native control parameters} &
\textbf{Timing and interface adaptations} \\
\midrule

Genie 3
& W/A/S/D translation and I/J/K/L view rotation
& Executed and recorded through the hosted interface;
maximum evaluated duration: 60\,s \\

HappyOyster
& W/A/S/D translation and I/J/K/L view rotation
& --- \\

LingBot-World-v2
& Translation: 0.06 units/frame;
rotation: $3.5^\circ$/frame
& Durations converted to frames at 16 FPS;
generation length padded to $4n+1$ frames \\

Matrix-Game 2.0
& 4 keyboard channels;
mouse increments: $\pm0.1$
& 6 control frames per 0.5\,s;
neutral padding or trimming to the required control length \\

Matrix-Game 3.0
& 6 keyboard channels;
mouse increments: $\pm0.1$
& 8 control frames per 0.5\,s;
57 initial generated frames, then 40 additional frames per block;
trimmed to the target length \\

Yume-1.5
& Structured control text:
\texttt{move=4.0}, \texttt{turn=4.0}
& Segment durations rounded to $29/16$\,s generation blocks,
with a minimum of one block \\

DreamX-World
& Translation: 0.075 units/frame;
rotation: $1.5^\circ$/frame
& Durations converted to frames at 16 FPS;
frame counts adjusted to latent-sequence length requirements;
output trimmed \\

ABot-World
& 8 keyboard channels;
combined inputs activate the corresponding keys jointly
& Each 0.5\,s benchmark action block mapped to an approximately
1\,s native generation block;
the full video uniformly resampled to the target frame count \\

Cosmos 3
& At a 10 FPS reference rate:
forward/backward: $+0.18/-0.10$ units/frame;
yaw: $3^\circ$/frame
& Increments scaled by $10/16$ at 16 FPS;
lateral inputs mapped to curved forward motion;
in-place turns approximated with forward motion;
pitch unsupported \\

Echo-WM
& Translation: 0.05 units/frame;
yaw/pitch: $0.4^\circ/0.2^\circ$/frame
& Durations converted to control frames at 24 FPS;
generation length follows $1+8n$;
neutral inputs retain inertial decay \\

Alaya-EVOKE
& Translation: 1.5 units/s;
rotation: $12^\circ$/s
& Camera poses constructed at 30 FPS for 24 FPS video;
generation blocks cover the target length,
followed by output trimming \\

Lyra 2.0
& Translation: 1.5 units/s;
rotation: $12^\circ$/s
& Camera trajectories constructed at 16 FPS;
output trimmed to the target length \\

\bottomrule
\end{tabularx}
\caption{
Native control parameters, timing conversions, and interface
adaptations for the evaluated models.
Translation values use each model's native coordinate units;
mouse increments and structured-text control values are native
interface parameters.
Rotation parameters are expressed in degrees where applicable.
In generation-length expressions, $n$ is an integer.
}
\label{tab:control-adapter-configs}
\end{table*}

\subsection{Evaluation Metrics}
\label{sec:video_metrics}

We evaluate generated videos along five complementary
dimensions: perceptual quality, temporal consistency,
revisit memory, physical and causal validity, and action
controllability.
Table~\ref{tab:metric_sampling} summarizes the evaluation
methods and input configurations.

\begin{table*}[tb]
\centering
\caption{
Evaluation methods and input configurations for video metrics.
Listed FPS values denote target sampling rates; native FPS
retains all decoded frames.
Spatial preprocessing preserves aspect ratio unless a fixed
size is specified.
A dash indicates an unspecified setting.
}
\label{tab:metric_sampling}

\begingroup
\fontsize{8.6pt}{10.0pt}\selectfont
\renewcommand{\arraystretch}{1.21}
\setlength{\tabcolsep}{5pt}
\setlength{\aboverulesep}{0pt}
\setlength{\belowrulesep}{0pt}

\resizebox{\textwidth}{!}{%
\begin{tabular}{@{}lllll@{}}
\toprule
\textbf{Dimension}
& \textbf{Metric}
& \textbf{Method / Backbone}
& \textbf{Temporal Sampling}
& \textbf{Spatial Preprocessing} \\
\midrule

\multirow{2}{*}{Perceptual Quality}
& Visual Quality (VQ)
& MUSIQ~\cite{ke2021musiqmultiscaleimagequality}
& 2 FPS
& Long side $\leq512$ px \\

& Dynamic Degree (DD)
& Farneb{\"a}ck flow~\cite{farneback2003two}
& 12 FPS
& Long side $\leq512$ px \\
\midrule

\multirow{3}{*}{Temporal Consistency}
& Jump-Based Continuity (JP)
& TransNetV2~\cite{soucek2024transnet}
& Native FPS
& $48\times27$ px ($W\times H$) \\

& Subject Consistency (SC)
& DINOv2-B/14 + CLIP-B/16
& 3 FPS
& DINO-S256; CLIP: - \\

& Background Consistency (BC)
& CLIP-B/32
& 2 FPS
& - \\
\midrule

\multirow{2}{*}{Revisit Memory}
& Global Memory Score (GMS)
& DINOv2-B/14
& Observation midpoints$^{a}$
& DINO-S256 \\

& Local Memory Score (LMS)
& DINOv2-B/14
& Observation midpoints$^{a}$
& DINO-S256 \\
\midrule

\multirow{3}{*}{Physical \& Causal Validity}
& Physical Causality (PC)
& Gemini 3.1 Pro Preview
& 8 global + interval samples$^{b}$
& Long side $\leq640$ px \\



& Content Causality (CC)
& Gemini 3.1 Pro Preview
& 8 global + interval samples$^{b}$
& Long side $\leq640$ px \\

& Evolution (EC)
& Gemini 3.1 Pro Preview
& 8 global + interval samples$^{b}$
& Long side $\leq640$ px \\
\midrule

\multirow{2}{*}{Action Controllability}
& Translation Response (TR)
& ViPE poses~\cite{huang2025vipevideoposeengine}
& Poses within action segments
& - \\

& Camera Rotation Response (CR)
& ViPE poses~\cite{huang2025vipevideoposeengine}
& Poses within action segments
& - \\
\bottomrule
\end{tabular}%
}

\vspace{3pt}
\begin{minipage}{\textwidth}
\fontsize{7.3pt}{8.8pt}\selectfont
\textit{Notes.}
DINO-S256 denotes resizing to a 256-pixel short side without
center cropping, followed by normalization and padding to
multiples of 14.
$^{a}$ The nearest valid frame to each observation-window
midpoint is selected.
$^{b}$ Eight uniformly spaced video-wide timestamps are
supplemented with each temporal interval's start, midpoint,
and end.
\end{minipage}

\endgroup
\end{table*}

\subsubsection{Perceptual Quality}
\label{sec:perception_metrics}

\paragraph{Visual Quality (VQ).}
VQ measures frame-level perceptual quality using
MUSIQ~\cite{ke2021musiqmultiscaleimagequality}.
The video score is the mean prediction over sampled frames,
capturing image quality without directly assessing temporal
coherence or instruction adherence.

\paragraph{Dynamic Degree (DD).}
DD measures motion intensity using Farneb{\"a}ck optical
flow~\cite{farneback2003two}.
For each adjacent sampled-frame pair, we remove the largest
1\% of flow magnitudes and average the remainder.
After normalization by the nominal sampling interval and
the image short-side length relative to 256 pixels,
the mean over frame pairs gives $d(v)$.
We define
\begin{equation}
    \mathrm{DD}(v)
    =100\sqrt{\min\!\left(1,\frac{d(v)}{d_{\mathrm{ref}}}\right)},
    \qquad d_{\mathrm{ref}}=102.889,
\end{equation}
where $d_{\mathrm{ref}}$ is the fixed 95th-percentile
reference from the calibration pool.
DD quantifies the magnitude of camera and scene motion.
Physical correctness is evaluated separately by PC.

\subsubsection{Temporal Consistency}
\label{sec:temporal_metrics}

\paragraph{Jump-Based Continuity (JP).}
JP measures abrupt discontinuities using frame-wise cut
probabilities from TransNetV2~\cite{soucek2024transnet}.
Predictions from overlapping temporal windows are averaged.
For the resulting probabilities $p_t$,
\begin{equation}
    \mathrm{JP}(v)=100\left(1-\max_t p_t\right).
\end{equation}
Higher scores indicate weaker evidence of a discontinuity.
The score is determined by the strongest predicted transition.

\paragraph{Subject Consistency (SC).}
SC combines adjacent-frame appearance stability with
retention relative to the initial frame.
Valid subject masks are used to gray out the background
and crop around the subject; otherwise, the full frame
is retained.
For $N$ sampled frames, let $d_i$ and $c_i$ denote DINOv2
ViT-B/14 features~\cite{oquab2024dinov2learningrobustvisual}
and CLIP ViT-B/16
features~\cite{radford2021learningtransferablevisualmodels}.
Using $s(x,y)=\max\{0,\cos(x,y)\}$,
\begin{equation}
    \mathrm{SC}(v)
    =\frac{100}{2(N-1)}
      \sum_{i=1}^{N-1}
      \left[s(d_{i-1},d_i)+s(c_0,c_i)\right].
\end{equation}
SC is evaluated on subject-mask-eligible cases.

\paragraph{Background Consistency (BC).}
BC applies the same adjacent-frame and initial-frame
comparison structure to CLIP ViT-B/32 background features
$b_i$:
\begin{equation}
    \mathrm{BC}(v)
    =\frac{100}{2(N-1)}
      \sum_{i=1}^{N-1}
      \left[s(b_{i-1},b_i)+s(b_0,b_i)\right].
\end{equation}
Subject regions are grayed out where masks are available;
otherwise, the full frame is used.
SC and BC assess subject and background appearance consistency across frames.

\subsubsection{Revisit Memory}
\label{sec:revisit_metrics}

Revisit memory evaluates appearance retention at predefined
observation anchors.
We extract L2-normalized, full-frame DINOv2 ViT-B/14 features
from the nearest valid frames to observation-window
midpoints.
For anchor $a$, let $z_{a,0}$ denote the initial observation
and $z_{a,1},\ldots,z_{a,K_a}$ its revisits.
Comparisons remain within the same anchor.

\paragraph{Global Memory Score (GMS).}
GMS compares each revisit with the initial observation:
\begin{equation}
    G_a
    =\frac{
        \sum_{k=1}^{K_a}w_{a,k}\,
        \bar{s}(z_{a,0},z_{a,k})
    }{
        \sum_{k=1}^{K_a}w_{a,k}
    },
    \qquad w_{a,k}=t_{a,k}-t_{a,0},
\end{equation}
where $t_{a,k}$ is the scheduled observation time and
$\bar{s}$ is cosine similarity clipped to $[0,1]$.
Elapsed-time weighting gives later revisits greater influence.

\paragraph{Local Memory Score (LMS).}
LMS compares successive revisits:
\begin{equation}
    L_a
    =\frac{
        \sum_{k=2}^{K_a}w_{a,k}\,
        \bar{s}(z_{a,k-1},z_{a,k})
    }{
        \sum_{k=2}^{K_a}w_{a,k}
    }.
\end{equation}
The initial-to-first-revisit pair is excluded to avoid
duplicating its contribution in GMS.
Both metrics use time since the initial observation as
the weight.
Anchor scores are averaged equally within each video;
LMS is unavailable when there are insufficient revisits.
Model-level aggregation uses complete cases with all
prescribed comparison pairs available.
These metrics assess visual retention at scheduled
checkpoints.

\subsubsection{Physical and Causal Validity}
\label{sec:physical_causal_metrics}

PC, CC, and EC use scene-specific checklists shared across
models for each case.
Gemini 3.1 Pro Preview evaluates each checklist item on a
1--5 scale, from complete failure to correct and natural
behavior, and provides supporting visual evidence.
Persistent constraints use video-wide observations;
temporal checks additionally request interval-specific
samples, as summarized in Table~\ref{tab:metric_sampling}.
Checks requiring change are judged by comparing states
over time, rather than by initial object presence.

\paragraph{Physical Causality (PC).}
PC evaluates whether scene contents obey applicable
physical constraints, including gravity, buoyancy,
collision response, deformation, material behavior, thermodynamics, and non-penetration of solid boundaries.
Checks distinguish persistent properties, such as support,
from event-specific responses, such as deformation
under impact.

\paragraph{Content Causality (CC).}
CC evaluates whether a scene-supported cause is accompanied
by a visually consistent effect on another object or region,
such as illumination producing corresponding shadows
or reflections.
The causal source is a scene object or natural phenomenon,
rather than the commanded camera or agent action.

\paragraph{Evolution Consistency (EC).}
EC evaluates spontaneous motion and state changes
independent of the commanded action sequence, such as
drifting, flowing, or melting.
It requires visible progression over the specified interval;
maintaining a static property does not constitute evolution.

\paragraph{Checklist examples.}

Table~\ref{tab:checklist-showcase} illustrates how the three
dimensions are instantiated as scene-specific checklist items,
with three examples per dimension. Each item is paired with
a supporting prompt excerpt.
The excerpts identify the scene content or events from which
the checks are derived; assessment concerns the physical
plausibility, causal consistency, and temporal evolution
of the rendered scene.
For example, PC assesses whether pedestrians' feet make plausible
contact with the ground, whereas EC assesses whether their gait
and movement continue over time. Similarly, the flag examples
distinguish material behavior under PC from attachment constraints
under CC. All items use the 1--5 scoring scale described above.

\begin{table*}[t]
\centering
\footnotesize
\setlength{\tabcolsep}{3.5pt}
\renewcommand{\arraystretch}{1.25}
\begin{tabular}{@{}llp{3.9cm}p{4.0cm}@{}}
\toprule
\textbf{Dim.} &
\textbf{Sub-type} &
\textbf{Checklist item} &
\textbf{Prompt evidence} \\
\midrule

\multirow{3}{*}{PC}
 & Gravity support
 & Do the boulders, gravel and glowing crystals rest stably on the ground under gravity?
 & ``moss-covered boulders \ldots a large glowing blue crystal'' \\

 & Rigid vs.\ soft
 & Do the flagpole, pavilion and fabric flags exhibit behavior appropriate to their materials?
 & ``gazebo \ldots picnic tables \ldots flags hanging partly unfurled'' \\

 & Contact exclusion
 & Do pedestrians' feet make physically plausible contact with the cobblestones while walking?
 & ``pedestrians walk across the ground'' \\

\midrule

\multirow{3}{*}{CC}
 & Light $\rightarrow$ reflection
 & Does the polished wood floor reflect the pendant lights consistently with the scene geometry and viewing angle?
 & ``The polished wood floor reflects a row of dome pendant lights'' \\

 & Constraint $\rightarrow$ attachment
 & Do the flags remain attached to the flagpole, with their motion constrained by the attachment?
 & ``flags hanging partly unfurled'' \\

 & Sun $\rightarrow$ specular
 & Are highlights on the ocean surface consistent with the setting sun and viewing angle?
 & ``waves crash \ldots against a setting sun over the ocean'' \\

\midrule

\multirow{3}{*}{EC}
 & Water $\rightarrow$ flow
 & Does the waterfall visibly continue flowing downward into the stream over 0--79\,s?
 & ``A waterfall flows down rocks \ldots into the stream'' \\

 & Steam $\rightarrow$ rise/drift
 & Does the steam visibly continue rising and drifting rightward over 0--90\,s?
 & ``steam plume rises \ldots drift horizontally rightward'' \\

 & Agents $\rightarrow$ locomotion
 & Do the pedestrians visibly continue walking across the square over 0--69\,s?
 & ``pedestrians walk across the ground'' \\

\bottomrule
\end{tabular}
\caption{
Representative checklist items for Physical Causality (PC),
Content Causality (CC) and Evolution Consistency (EC).
Prompt excerpts (rightmost column) provide the scene context
from which the checks are derived.
Each item assesses whether the \emph{rendered} scene exhibits
physically plausible behavior, preserves causal dependencies
or evolves consistently over time, and is scored using the
1--5 scale described in the text.
Assessment is based on observations in which the relevant
objects or effects can be evaluated.
}
\label{tab:checklist-showcase}
\end{table*}

\subsubsection{Action Controllability}
\label{sec:control_metrics}

We evaluate camera-action following using trajectories
recovered by ViPE~\cite{huang2025vipevideoposeengine}.
Each non-idle action segment is assessed using its first
and last available camera-to-world poses,
$(R_s,t_s)$ and $(R_e,t_e)$.

\paragraph{Translation Response (TR).}
TR compares the commanded direction with displacement
in the segment's initial camera coordinates:
\begin{equation}
    \Delta t_{\mathrm{cam}}=R_s^\top(t_e-t_s).
\end{equation}
Forward, backward, left, and right correspond to
$+z$, $-z$, $-x$, and $+x$; combined commands sum their
direction vectors.
A segment is correct when the recovered and expected
directions differ by at most $60^\circ$.
Zero or undefined displacements do not count as correct.

\paragraph{Camera Rotation Response (CR).}
CR evaluates yaw and pitch extracted from $R_s^\top R_e$.
A commanded rotation is correct when its sign matches
the instruction and its absolute magnitude is at least
$4^\circ$.
Commands involving both axes must satisfy both conditions.

For correct counts $C_T,C_R$ and eligible segment counts
$N_T,N_R$,
\begin{equation}
    \mathrm{TR}(v)=100\frac{C_T}{N_T},
    \qquad
    \mathrm{CR}(v)=100\frac{C_R}{N_R}.
\end{equation}
Combined translation--rotation segments contribute to
both metrics.
Segments with insufficient pose coverage remain in the
denominator, while segments entirely within an identified
corrupt prefix are excluded.
Model scores average eligible videos equally.
TR and CR quantify directional agreement between the prescribed controls and the recovered camera motion.

\paragraph{Aggregation Protocol.}
The main table reports video-equal macro averages, whereas
Figure~\ref{fig:actiontype} pools valid action segments across videos
within each action type (micro averaging). Consequently, the two
results may differ because videos contain different numbers of action
segments. For example, DreamX-World obtains a pooled rotation accuracy
of $89.04\%$ ($6497/7297$), compared with the video-equal score of
$90.32\%$ in Table~\ref{tab:main_results}.

\section{Audio Evaluation}
\label{app:audio_evaluation}

We evaluate audio generated by Echo-WM and HappyOyster using five
complementary measures: Audio Quality (AQ),
ImageBind~\citep{girdhar2023imagebind} similarity (IB),
Synchformer~\citep{iashin2024synchformer} offset (SF),
Audio Realism (AR), and Action--Audio Correspondence (AAC).
These measures characterize the sound itself and its
relationship with visual content.
Gemini 3.1 Pro Preview serves as the judge for AR and AAC.

\paragraph{Audio Quality (AQ).}
Audiobox Aesthetics~\citep{tjandra2025audioboxaesthetics}
evaluates the complete audio track and predicts Production Quality (PQ) and Content
Usefulness (CU).
We define a composite score as their unweighted mean:
\begin{equation}
    \mathrm{AQ}_v
    =\frac{\mathrm{PQ}_v+\mathrm{CU}_v}{2}.
\end{equation}

\paragraph{ImageBind Similarity (IB).}
IB measures audio--visual semantic similarity.
Within annotated non-idle action intervals, we exclude
0.25 seconds at each boundary and sample six-second
windows where sufficient duration remains.
Each window pairs its audio with the midpoint video frame.
For audio and visual embeddings $\mathbf{a}_i$ and
$\mathbf{v}_i$, the window score is
\begin{equation}
    c_i
    =\frac{\mathbf{a}_i^\top\mathbf{v}_i}
           {\|\mathbf{a}_i\|_2\|\mathbf{v}_i\|_2}.
\end{equation}
The video score averages valid windows using newly covered
duration as the weight, reducing the influence of an
overlapping final window.

\paragraph{Synchformer Offset (SF).}
SF estimates audio--visual temporal offset in five-second
windows using the same action-interval boundary rule.
For predicted offset $\Delta_i$ and newly covered duration
$w_i$, the video-level mean absolute offset is
\begin{equation}
    \mathrm{SF}_v
    =\frac{\sum_{i\in\mathcal{W}_v}w_i|\Delta_i|}
           {\sum_{i\in\mathcal{W}_v}w_i},
\end{equation}
where $\mathcal{W}_v$ contains valid windows.
SF is treated as exploratory because its predictions
did not reliably recover known synthetic shifts
during validation.

\paragraph{Audio Realism (AR).}
We divide each video into consecutive segments of up to
20 seconds.
Gemini 3.1 Pro Preview evaluates each segment using audio
alone, assessing acoustic texture, continuity, and signal
integrity on a 1--5 scale.
AR measures acoustic realism, while AAC evaluates
correspondence with visible actions.

\paragraph{Action--Audio Correspondence (AAC).}
Within the same segments, Gemini 3.1 Pro Preview independently
describes audible events in four-second audio windows and
visible actions in matching windows of a silent video
sampled at 2 FPS.
The same model then compares these descriptions and assigns
a 1--5 score when sufficient evidence is available.
AAC measures whether identifiable actions have appropriate
sound responses and whether their changes correspond within
the same temporal window.

For $m\in\{\mathrm{AR},\mathrm{AAC}\}$, segment scores are
aggregated by duration:
\begin{equation}
    m_v
    =\frac{
        \sum_{j\in\mathcal{S}_v^m}\ell_j s_j^m
    }{
        \sum_{j\in\mathcal{S}_v^m}\ell_j
    },
\end{equation}
where $\mathcal{S}_v^m$ contains scored segments,
$s_j^m$ is the segment rating, and $\ell_j$ is its duration.
For all window- and segment-based audio metrics, aggregation
uses only valid scores.
Videos with no valid observations receive no score,
and temporal coverage is reported alongside the averages.

\section{Agreement with Human Model Rankings}
\label{app:human_model_rankings}
For the ranking experiment in Table~\ref{tab:human_model_rankings}, three
independent annotators scored videos from ten open-weight models on 40 tasks
stratified by benchmark suite and target duration.
For each metric, let $x_m$ and $y_m$ denote the
automatic and human scores aggregated over the same
evaluated videos for model $m$. Each correlation is
computed across $M=10$ models.

\paragraph{Spearman rank correlation (SRCC).}
Let $r_m$ and $q_m$ denote the ranks of $x_m$ and
$y_m$, respectively. Model-ranking agreement is computed as

\begin{equation}
\mathrm{SRCC}
=
\frac{
\sum_{m=1}^{M}(r_m-\bar r)(q_m-\bar q)
}{
\sqrt{
\sum_{m=1}^{M}(r_m-\bar r)^2
\sum_{m=1}^{M}(q_m-\bar q)^2
}
}.
\end{equation}

\paragraph{Pearson linear correlation (PLCC).}
We apply a VQEG five-parameter logistic mapping
to the automatic scores, yielding
$\tilde{x}_m=g_{\boldsymbol{\beta}}(x_m)$ on the
human-score scale. Linear agreement is then computed as

\begin{equation}
\mathrm{PLCC}
=
\frac{
\sum_{m=1}^{M}
(\tilde{x}_m-\bar{\tilde{x}})(y_m-\bar y)
}{
\sqrt{
\sum_{m=1}^{M}(\tilde{x}_m-\bar{\tilde{x}})^2
\sum_{m=1}^{M}(y_m-\bar y)^2
}
}.
\end{equation}

SRCC summarizes the ordering of the ten models. PLCC summarizes linear
agreement after mapping their automatic scores to the human-score scale.

For Figure~\ref{fig:judge_ranking}, each judge's model score is the unweighted
mean of PC, CC, and EC. We compare its ordering of the ten models with the
ordering from the three annotators' mean scores using Kendall's $\tau$ and
Spearman's $\rho$. The human reference in the figure averages the correlations
from the three annotator pairs.

\section{Qualitative Case Studies}
\label{app:qualitative}

\begin{figure*}[p]
    \centering
    \includegraphics[
        width=\textwidth,
        height=\textheight,
        keepaspectratio
    ]{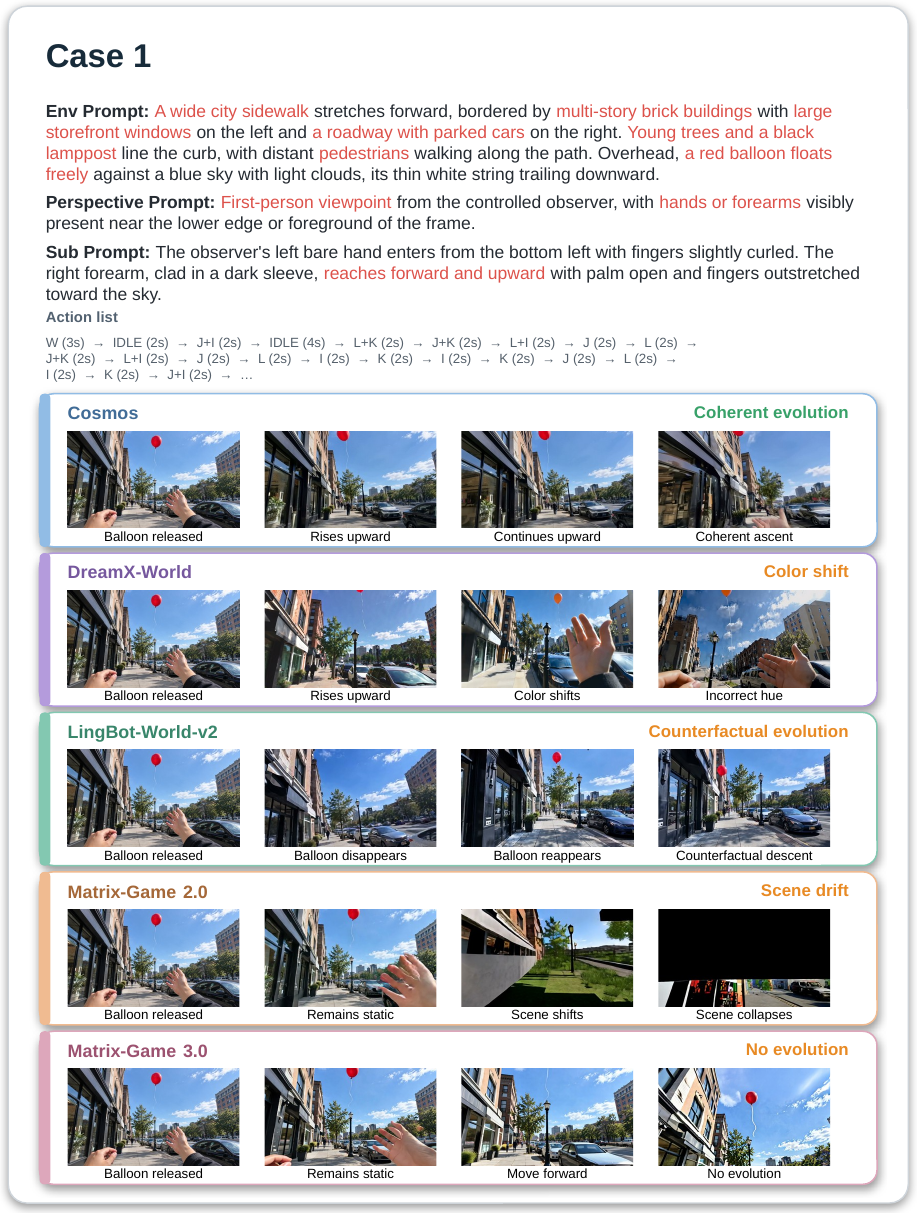}
    \caption{
    Ongoing scene evolution during first-person navigation. Frames show
    balloon ascent, appearance changes, and scene drift at selected times for each model.
    Cosmos~3 shows coherent balloon ascent, while other examples
    exhibit color changes, reappearance after disappearance,
    scene drift, or little visible progression.
}
    \label{fig:case-evolution}
\end{figure*}

\begin{figure*}[p]
    \centering
    \includegraphics[
        width=\textwidth,
        height=\textheight,
        keepaspectratio
    ]{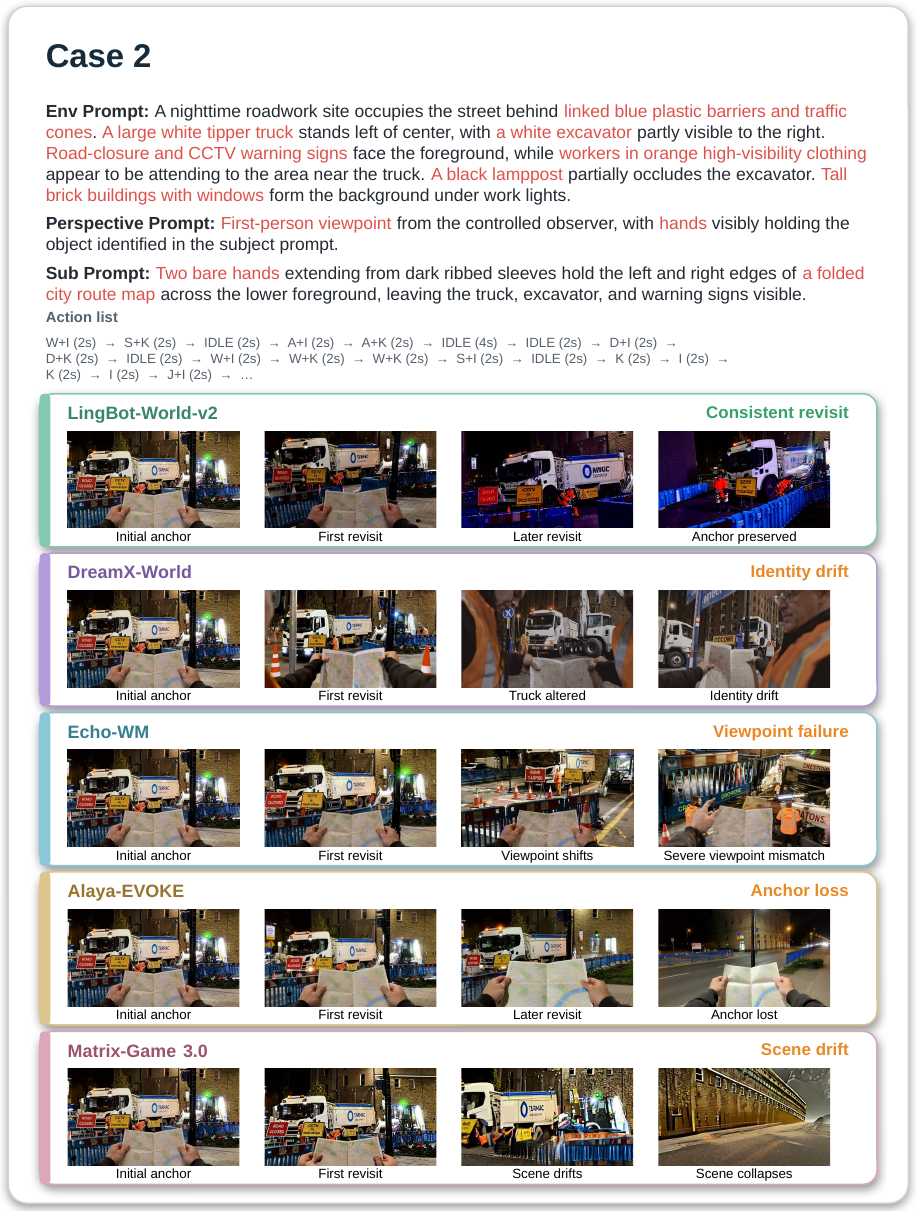}
    \caption{
    Qualitative comparison on world persistence across repeated revisits.
    LingBot-World-v2 preserves the initial anchor most consistently,
    whereas other models exhibit identity drift, viewpoint failure,
    anchor loss, or scene drift.
    }
    \label{fig:case-revisit}
\end{figure*}

\begin{figure*}[p]
    \centering
    \includegraphics[
        width=\textwidth,
        height=\textheight,
        keepaspectratio
    ]{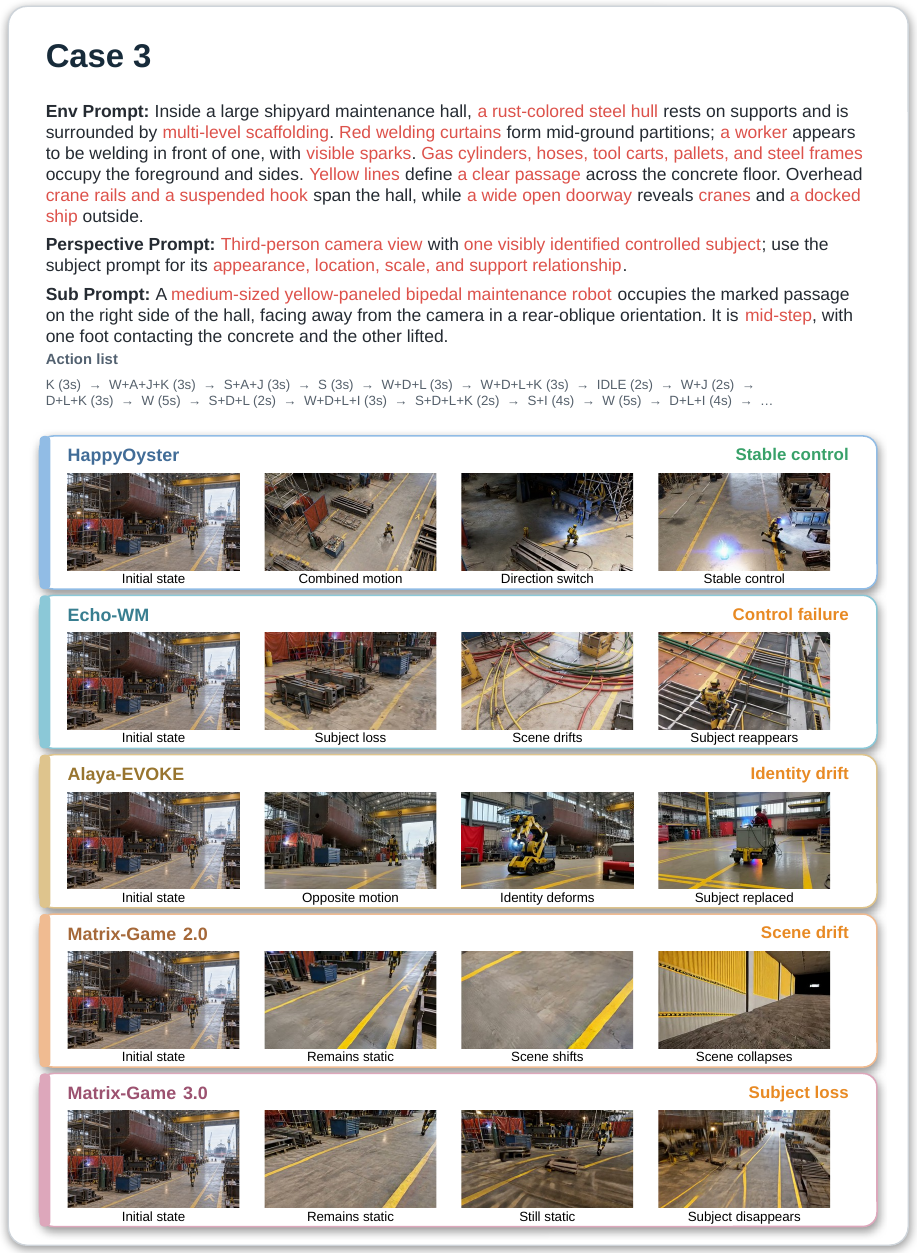}
    \caption{
    Sustained control under complex action sequences. Each column shows frames
    selected for the robot's motion, appearance, or loss from view at model-specific
    action stages. HappyOyster exhibits coherent subject motion in the displayed frames, while other models exhibit control failure, identity drift, scene drift, or subject loss.
    }
    \label{fig:case-control}
\end{figure*}

The 80-second street sequence follows a floating red balloon during first-person
navigation (Figure~\ref{fig:case-evolution}). The nighttime roadwork sequence
returns repeatedly to the same anchor (Figure~\ref{fig:case-revisit}).
In the shipyard, a maintenance robot receives combined commands and direction
changes (Figure~\ref{fig:case-control}).

\clearpage

\clearpage
\section{Real-World Video References}
\label{sec:real_world_cases}
\begin{figure}[p]
\centering
\includegraphics[width=\linewidth,height=0.87\textheight,keepaspectratio]{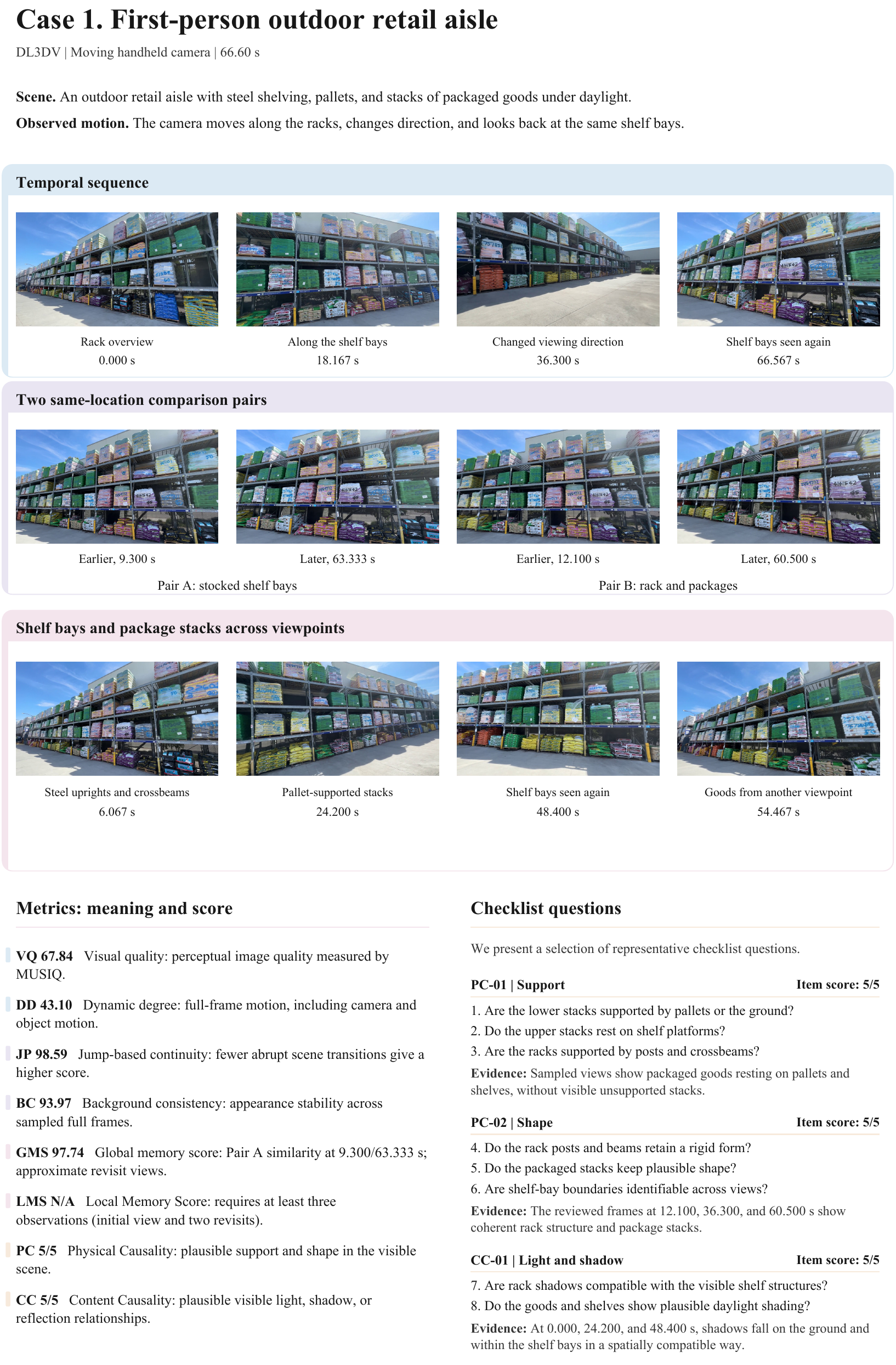}
\caption{DL3DV outdoor retail aisle. A handheld camera passes shelves and revisits two shelf bays.}
\label{fig:real_world_ego}
\end{figure}

\begin{figure}[p]
\centering
\includegraphics[width=\linewidth,height=0.87\textheight,keepaspectratio]{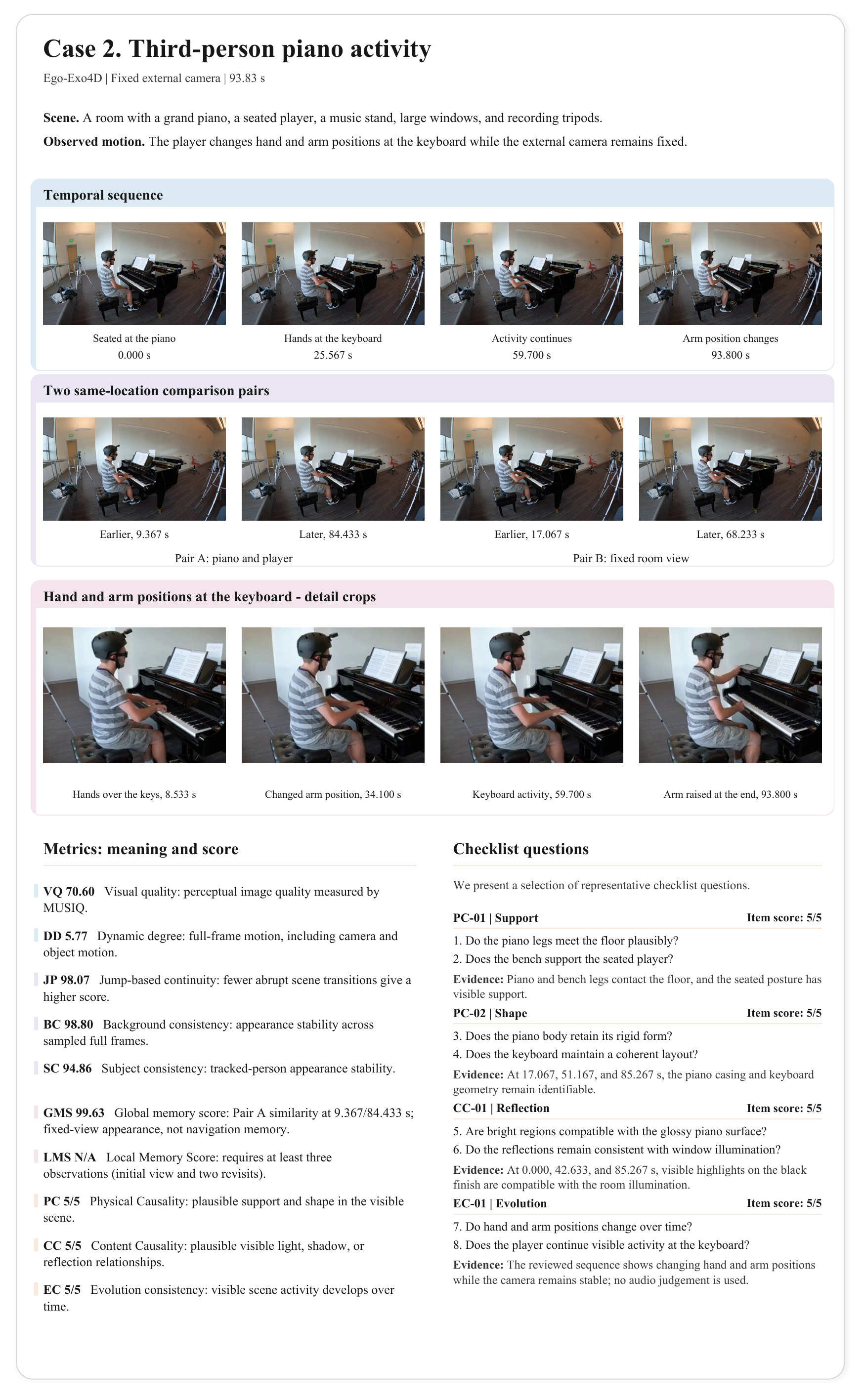}
\caption{Ego-Exo4D piano sequence. A fixed camera records the pianist's hands and arms across the performance.}
\label{fig:real_world_exo}
\end{figure}

The two sequences in Figures~\ref{fig:real_world_ego} and~\ref{fig:real_world_exo}
show how the metrics score a moving retail-aisle view and a fixed view of a pianist.
\subsection{Real-Video Reference Evaluation}
\label{app:real_video}

To complement the two case studies, we report aggregate
results on 80 real videos totaling approximately 93 minutes:
50 egocentric DL3DV-10K recordings of indoor and outdoor
exploration (56--94 seconds) and 30 exocentric Ego-Exo4D
recordings (63--120 seconds), spanning 24 locations and
seven activities, including basketball, soccer, climbing,
cooking, and bicycle repair.
Table~\ref{tab:real_video_metrics} reports equally weighted
video-level means on the 0--100 scale.
TR/CR are unavailable because these recordings lack independently
prescribed control-command sequences.

The largest gaps between the real-video references and generated videos appear in shot continuity and background consistency (Tables~\ref{tab:main_results} and~\ref{tab:real_video_metrics}).
Both real-video sets score about 98 on continuity (JP) and above 92 on background consistency (BC), compared with generated-model maxima of 88.87 and 86.91, respectively.
The gap in frame-level visual quality is smaller. Mean VQ scores for the two real-video sets are 68.95 and 70.19, compared with a maximum of 68.61 for the generated videos.
Viewpoint changes, articulation, and occlusion in real footage can also lower cross-frame feature similarity.

\begin{table}[H]
\centering
\caption{
Real-video reference scores on a 0--100 scale.
Values are equally weighted video-level means. Each anchor has two observations, so LMS-2obs equals GMS.
}
\label{tab:real_video_metrics}
\begingroup
\small
\setlength{\tabcolsep}{4pt}
\renewcommand{\arraystretch}{1.05}

\begin{tabular*}{\linewidth}
{@{\extracolsep{\fill}}lccccccc@{}}
\toprule
Real-video set & VQ & DD & JP & SC & BC & GMS & LMS-2obs \\
\midrule
Egocentric, 50 videos
& 70.19 & 43.49 & 98.09 & — & 92.43 & 85.30 & 85.30 \\
Exocentric, 30 videos
& 68.95 & 11.75 & 97.94 & 84.10 & 95.77 & 88.24 & 88.24 \\
\bottomrule
\end{tabular*}

\endgroup
\end{table}

\paragraph{Motion and consistency settings.}

DD measures motion intensity: largely stationary
exocentric cameras yield 11.75, while egocentric camera
movement yields 43.49. SC and BC are sampled at approximately 1 FPS.
Egocentric SC is omitted because persistent foreground
subjects and their masks are not annotated; exocentric SC
uses masks of the main activity participants.

\paragraph{Two-observation appearance comparisons.}

For egocentric videos, reconstructed camera poses identify
earlier and later observations with similar positions and
viewing directions. For exocentric videos, frames at
approximately 10\% and 90\% of each recording provide
fixed-view cross-time comparisons.
Because these recordings provide two annotated observations
per anchor, we report LMS-2obs, which includes the
initial-to-first-subsequent-observation pair.
It equals GMS with two observations by construction.
Exocentric comparisons assess appearance retention from
a fixed view. The generated-video benchmark uses its prescribed revisit anchors.
\paragraph{Supplementary checklist assessment.}

We additionally evaluate scene-specific PC, CC, and EC
checklists on the same 80 real videos, covering 329 checklist
questions in total. Both the 50-video egocentric set and the
30-video exocentric set score 100 on each of PC, CC, and EC
on the 0--100 scale, where 100 corresponds to a rating of 5/5.
The sampled evidence covers support, material behavior, shape, lighting,
and ongoing scene activity.

\clearpage
\section{Prompts}
\label{app:prompt}
\subsection{Prompt Rewriting and Verification}

To improve the descriptive completeness of the initial-frame prompts,
we use a multi-model rewriting and verification procedure involving
GPT-5.6 Sol~\citep{openai2026gpt56sol},
Gemini 3.8 Flash~\citep{google2026gemini38flash},
Claude Sonnet 5~\citep{anthropic2026claudesonnet5}, and
Qwen3.8-Max~\citep{alibaba2026qwen38max}.
For each case, the initial image and the original perspective,
environment, and subject descriptions are provided as input.
Two models independently rewrite the environment and subject fields,
while a third model verifies the candidates against the image and
source descriptions. Model assignments follow a balanced fixed
rotation across data splits, source origins, viewpoints, and subject
presentations. Candidate selection prioritizes source-fact
preservation, visual grounding, spatial relations, and subject
consistency. The perspective description remains unchanged, and all
candidate outputs, verification decisions, and model identifiers are
retained for auditing.

We apply this procedure uniformly to 300 cases, with the four models receiving approximately equal numbers of candidate-generation and verification assignments. After rewriting, the median combined length of the environment and subject descriptions increases from approximately 47 to 108 words, with the environment description increasing from 25 to 75 words and the subject description from 22 to 35 words. The rewritten prompts provide richer descriptions of stable landmarks, spatial layout, foreground–background relations, occlusions, support and contact relations, and the observable states of dynamic objects. This improves the completeness and visual verifiability of the initial scene descriptions for long-horizon navigation, state persistence, revisit consistency, and interactive behavior evaluation.

\begin{rewritepromptbox}[Image-grounded rewriting prompt]
You are an image-grounded prompt writer for a long-horizon interactive world-model benchmark.

Given the initial image and the source perspective, environment, and subject descriptions,
rewrite only the environment and subject fields. Keep the perspective description unchanged.

Preserve important source facts and add only details supported by the image. Describe stable
landmarks, spatial layout, foreground/background relations, paths, openings, occlusions,
support, contact, and visible object states.

For potentially dynamic elements, describe their visible current state. Describe an apparent
activity only when supported by visual evidence such as pose, motion blur, displacement,
flowing, or waving. Do not infer hidden intentions, goals, or future actions. A car on a road
is not necessarily driving, and a person near a table is not necessarily eating.

For first-person views, describe only visible hands, arms, held objects, or interfaces. For
third-person views, describe only the visible controlled subject. Do not invent unseen body
parts, objects, identities, text, exact counts, hidden spaces, or unsupported functionality.
Avoid poetic or cinematic filler.

Return JSON only in the following format:

\begin{verbatim}
{
  "rewrite_environment_prompt_en": "...",
  "rewrite_subject_prompt_en": "..."
}
\end{verbatim}

Use soft length targets of approximately 60--85 words for the environment and
20--40 words for the subject.
\end{rewritepromptbox}

\begin{verifypromptbox}[Independent verification prompt]
You are an independent visual-grounding verifier, not a stylistic editor.

Given the initial image, the source descriptions, and two rewrite candidates, evaluate each
candidate independently. Check whether each candidate preserves important source facts,
adds only image-supported details, maintains spatial relations and viewpoint constraints,
and describes current activities only when visually supported.

Mark a candidate as FAIL if it contains a material hallucination, an incorrect spatial or
contact relation, an unsupported activity, a viewpoint or visible-body inconsistency, or an
important omitted source fact. Do not reward verbosity and do not rewrite or paraphrase the
candidates.

If exactly one candidate passes, select it. If both pass, select the one with fewer unsupported
additions; if they are equivalent, select Candidate A. If both fail, return \texttt{NONE} and
request fallback generation.

Return JSON only in the following format:

\begin{verbatim}
{
  "candidate_a": {
    "status": "PASS",
    "issues": []
  },
  "candidate_b": {
    "status": "FAIL",
    "issues": [
      "unsupported activity"
    ]
  },
  "selected": "A",
  "fallback_required": false
}
\end{verbatim}
\end{verifypromptbox}

\subsection{Physical and Causal Validity Evaluation Prompts}
\label{app:physical_causal_prompts}

We use a shared two-stage prompting pipeline for
\textbf{Physical Causality (PC)}, \textbf{Content Causality (CC)}, and
\textbf{Evolution Consistency (EC)}. The first prompt extracts a
scene-specific checklist from the initial-frame descriptions and the timed
action sequence. The second prompt evaluates the generated video against the
applicable checklist items. The complete prompts used in both stages are provided below to specify the
evaluation criteria, scoring rules, and required output formats.


\begin{rewritepromptbox}[PC/CC/EC checklist extraction prompt]
You are a checklist extraction expert for world-model evaluation.

You are given a case description consisting of environment, subject, and
perspective prompts, which describe the static first frame, together with a
timestamped action sequence describing camera or observer movement. Extract
checklist items for three dimensions: Physical Causality (PC), Content
Causality (CC), and Evolution Consistency (EC). All items will be evaluated
on a 1--5 scale.

\textbf{Core principle.} This is an image-to-video setting. The generated
video should animate the static scene while preserving physically correct
behavior. The checklist should capture the physical laws and natural dynamics
that the scene should exhibit. Infer these requirements from physical common
sense, not only from explicit motion verbs in the prompt. For example, a
floating balloon should drift or rise, and visible water should exhibit
physically plausible water behavior.

\textbf{Step 1: determine applicability.}

\begin{itemize}[leftmargin=1.4em,itemsep=2pt,topsep=3pt]
    \item \textbf{PC} is applicable when the scene contains objects or
    phenomena governed by physical constraints, including gravity and support,
    buoyancy, collision, deformation, material behavior, thermodynamics,
    fluids, soft bodies, smoke, or fire. A scene containing ordinary physical
    objects is usually applicable because gravity, support, and material
    behavior can be checked.

    \item \textbf{CC} is applicable when the scene supports a cross-object
    causal relation in which a scene object or natural event causes an effect
    on another object or region. Examples include light producing shadows or
    reflections, wind moving an object, fire igniting material, or water
    making a surface wet. Scene-supported physical inference is allowed, but
    do not invent a causal source absent from the prompt or image.

    \item \textbf{EC} is applicable when the scene contains an object or
    phenomenon that should move or change state spontaneously, such as a
    drifting balloon, flowing or rippling water, moving clouds, walking
    people, wind-driven flags or plants, dispersing smoke or fog, melting ice,
    spreading fire, or growth. Decide based on whether the object would
    naturally change in the physical world, not only on whether the prompt
    contains a motion verb. Treat an explicitly static state such as calm,
    static, still, moored, or seated as non-evolving when appropriate.
    Exclude motion directly driven by the action sequence, because commanded
    camera or subject motion is not spontaneous evolution.
\end{itemize}

\textbf{Step 2: extract items only for dimensions with
\texttt{applicable=true}.}

\textbf{1. Physical Causality (PC).}
Each item must express one physical law as a positive assertion of how an
object should behave; do not write conditional ``if--then'' statements.
Allowed subtypes are \texttt{gravity}, \texttt{buoyancy},
\texttt{collision}, \texttt{deformation}, \texttt{material},
\texttt{thermodynamics}, and \texttt{impenetrability}.
Extract one item for each applicable subtype, normally three to six items.

Persistent laws such as gravity, buoyancy, material behavior, and
thermodynamics use \texttt{temporal=null}. Event-specific collision,
deformation, and impenetrability checks use a time interval derived from
the action sequence.

Use \texttt{deformation} only for event-driven deformation. Static draping,
bending, rigid shape preservation, and similar persistent properties belong
to \texttt{material}.

For \texttt{impenetrability}, extract at most one item when the camera,
controlled subject, vehicle, or a rigid attached object moves near a solid
boundary. Check that it does not pass through an intact wall, floor, terrain,
closed door, fence, furniture item, or other solid object. Ordinary occlusion,
valid contact, stopping, detouring, and passage through a genuine opening are
not penetration.

Examples include: ``The controlled subject should remain outside the intact
wall while approaching it,'' and ``The first-person viewpoint should pass
through the open doorway rather than through the adjacent wall.''

\textbf{2. Content Causality (CC).}
Each item must express one cross-object causal chain in the form
``because A occurs, B should appear or change.'' The cause A must be a scene
object or natural event, not an agent action. Extract approximately three
items when supported. Every CC item must include a temporal interval. A
continuously active relation may use the full video interval.

\textbf{3. Evolution Consistency (EC).}
Each item must express one spontaneous process that changes over time,
either \texttt{motion\_change} or \texttt{state\_change}, and must include a
temporal interval. Write a positive assertion specifying how the element
should gradually or continuously change between \texttt{since} and
\texttt{until}. Extract one or two items per applicable subtype, normally
two to four items.

A persistent state is not an evolution: continuously glowing, remaining
inflated, remaining upright, or remaining still should not be extracted
as EC.

\textbf{Hallucination policy.}
Do not introduce objects or events absent from the scene, such as an
unsupported fire, jump, object twitch, or traffic flow. Do not contradict
an explicitly static description, such as requiring ripples on water
described as calm. Standard physical consequences of visible scene elements
are allowed, including balloon drift, water motion, cloud drift, cloth
draping, and light-induced shadows.

Return strict JSON without Markdown:

\begin{verbatim}
{
  "case_id": "...",
  "applicability": {
    "PC": {"applicable": true, "reason": "...", "cues": []},
    "CC": {"applicable": false, "reason": "...", "cues": []},
    "EC": {"applicable": true, "reason": "...", "cues": []}
  },
  "items": [
    {
      "id": "PC-01",
      "dimension": "PC",
      "metric_key": "gravity",
      "check_type": "physics",
      "question": "...",
      "verify": "...",
      "expected": "...",
      "source": "prompt|inferred",
      "temporal": null,
      "required": true
    }
  ]
}
\end{verbatim}

The allowed \texttt{metric\_key} values are:

\begin{verbatim}
PC: gravity | buoyancy | collision | deformation |
    material | thermodynamics | impenetrability
CC: cross_object_causal
EC: motion_change | state_change
\end{verbatim}

The \texttt{temporal} field must be either \texttt{null} or an object
of the form \texttt{\{"since":"...","until":"..."\}}. All EC items,
all CC items, and event-specific PC collision, deformation, or
impenetrability items require a temporal object. Persistent PC gravity,
buoyancy, material, and thermodynamics items use
\texttt{temporal=null}. Do not output items for a dimension marked
\texttt{applicable=false}.
\end{rewritepromptbox}

\begin{verifypromptbox}[PC/CC/EC video judging prompt]
You are a world-model video-evaluation judge. You are given sampled video
frames, each labelled with a timestamp, and a set of checklist items.
Evaluate every item and return an integer score from 1 to 5 together with
one sentence of visual evidence.

Use the following scale:

\begin{verbatim}
5 = completely correct and natural
4 = mostly correct, with only a very minor defect
3 = a clear but non-severe problem
2 = a severe problem
1 = completely incorrect or corrupted
\end{verbatim}

\textbf{Temporal evaluation is essential.}

For \texttt{temporal=null}, evaluate representative frames across the full
video and determine whether the persistent property, such as gravity,
support, or material behavior, holds throughout.

For \texttt{temporal=\{since,until\}}, compare the beginning and ending
states of the specified interval. If an item requires an object to change
over time, such as drifting, melting, moving, or otherwise evolving,
verify that the ending state actually differs in the required way. The
object's correct presence in the initial frame is not sufficient.

If the required change is absent at the end of the interval, assign a score
of 1 or 2. For example, if an item requires a balloon to rise and drift away
from 0 to 80 seconds, but the balloon remains at its initial location in
the final interval frame, assign a score of 1 or 2.

Use only visible evidence from the supplied frames. Do not infer unsupported
events. If the critical evidence is unclear, score conservatively. Ignore
ordinary changes caused by viewpoint, illumination, or occlusion, and
evaluate only the physical, causal, or evolutionary phenomenon named by
the item.

Return strict JSON without Markdown:

\begin{verbatim}
{
  "verdicts": [
    {"id": "EC-01", "score": 4, "evidence": "..."}
  ]
}
\end{verbatim}

Every \texttt{score} must be an integer from 1 to 5.
\end{verifypromptbox}

\subsection{Audio Evaluation Prompts}
\label{app:audio_prompts}

We use four prompts to support the audio evaluation pipeline, covering
\textbf{Audio Realism (AR)} and
\textbf{Action--Audio Correspondence (AAC)}.

AR evaluates the perceptual realism and recording fidelity of the generated
audio from the audio track alone, focusing on generation artifacts, temporal
continuity, and signal integrity. The AR judge directly inspects the complete
audio recording and assigns an integer score from 1 to 5.

AAC evaluates whether visible actions are accompanied by plausible audible
responses and whether their coarse temporal changes are consistent. To reduce
cross-modal bias, AAC is implemented in three stages. An
\textbf{audio-only event observer} independently identifies acoustic events
within labelled temporal windows, while a
\textbf{silent-video action observer} independently describes visible actions,
motion changes, and potentially sound-producing events from the corresponding
video windows. An \textbf{evidence-linked AAC judge} then compares the two
observation streams and scores their correspondence based only on supported
cross-modal evidence.

This separation prevents the audio observer from inferring actions from visual
context and prevents the visual observer from imagining sounds. The final AAC
judge therefore operates on independently extracted evidence rather than on the
raw multimodal input. The complete prompts used for each stage are provided
below.

\begin{rewritepromptbox}[Audio Realism (AR) Judge]
You are an audio-only judge of recording realism for AI-generated
world-model audio. Evaluate audible generation artifacts and recording
fidelity from the supplied audio alone. Judge the received sound, not an
imagined video or action instruction.

Inspect the entire supplied recording for the following three types of evidence
before assigning one Audio Realism (AR) score:

\textbf{1. Texture and transients.}
Identify unnatural metallic or robotic coloration in voices and environmental
sounds; phasey, hollow, or warbling textures; smeared impacts; or severe
muffling or bandwidth degradation audible within this recording.

\textbf{2. Continuity.}
Identify artificial clicks, bursts, dropouts, splice or loop seams, or a noise
floor that pumps or changes abruptly. Regular footsteps, flowing water, and
repeated actions are not defects merely because they repeat. If alleging a
loop, identify the audible seam or unnatural repeated texture.

\textbf{3. Signal integrity.}
Identify clipping, harsh distortion, intrusive electronic interference, or
collapse into unstable or mechanical noise. Distinguish natural wind, water,
and friction from electronic artifacts using their audible texture, not a
guessed scene.

\textbf{World-model adaptation.}
Common content includes locomotion, breathing, clothing or prop friction,
water, and environmental ambience. It may also contain speech or music.
Do not assume any of these must occur. Do not score whether a footstep matches
a visible action or material, whether an environment is appropriate, or whether
a voice has a visible speaker; those require other metrics.

Do not excuse a distorted voice, stutter, or metallic texture merely by calling
it an intentional vocoder, glitch effect, or sound design. Intention is not available from the input. Conversely, an electronic instrument or musical
passage is not automatically defective. Describe the actual evidence and state
uncertainty where legitimate musical timbre and a generation defect cannot be
distinguished.

The common evaluation format is 24~kHz mono. Do not penalize mono presentation
or assume lost frequencies above its available bandwidth. Do not require studio
silence; natural recording noise and quiet scenes can be realistic.

For a long recording, inspect the beginning, middle, and end, as well as
any audible transitions. A clean majority does not erase a clearly
defective passage. Report approximate regions or times only when supported,
and do not claim coverage of intervals that were not reliably inspected.

\textbf{Scoring: integer 1--5.}

\textbf{1:} Pervasive severe corruption or artificial texture makes most
meaningful sound difficult to recognize.

\textbf{2:} Pronounced, sustained, or recurring mechanical coloration,
distortion, muffling, or unstable noise substantially damages recording realism.

\textbf{3:} Sound remains intelligible, but identifiable generation-like
texture, phase effects, noise, or continuity defects are clearly noticeable in
relevant passages.

\textbf{4:} Convincing recording quality overall; remaining imperfections are
small and isolated, such as a slightly softened impact or faint brief artifact.

\textbf{5:} Convincingly recorded textures, resolved transients, and stable
signal continuity throughout the supplied recording, with no identifiable
generation artifact. Support this with concrete audible observations;
clear or intelligible sound alone is insufficient.

Choose the anchor supported by severity, recurrence, and temporal extent.
Do not force a distribution, a target mean, or a low score. Do not replace
uncertainty with invented defects. No N/A. A score of 5 must not coexist with
acknowledged clear unresolved defects in the reason.

Return JSON only. Keep each observation brief and evidence-based; do not provide
hidden reasoning.

\begin{verbatim}
{
  "texture_observation": "...",
  "continuity_observation": "...",
  "signal_observation": "...",
  "defect_regions": [
    {
      "region": "approximate portion of the recording",
      "evidence": "audible property, not presumed intent"
    }
  ],
  "uncertainties": [],
  "reason": "brief justification of the selected anchor",
  "AR": 1
}
\end{verbatim}
\end{rewritepromptbox}

\begin{rewritepromptbox}[AAC Audio-only Event Observer]
You are an audio-only event observer. Listen to the labelled, successive
audio clips and process \textbf{each clip independently}. No video or scene
description is available.

Report acoustic events in that clip and split distinct sounds into distinct
events. An impact, an animal-like vocalization, and steady ambience must be
separate entries. A source label attached to one event cannot be copied onto
another event merely because they co-occur.

For each event, report:

\begin{itemize}
    \item its acoustic description;
    \item its event type: impact, vocalization, ambience, speech, music, or other;
    \item a specific source only if it is recognizable from that event alone.
\end{itemize}

In particular, rhythmic thuds do not prove a horse, a gallop, human walking,
or any contact material. A broadband rush does not prove wind or water.
When identity is uncertain, write \texttt{unresolved} and describe the sound
itself instead.

Contact material is \texttt{unresolved} unless it is independently audible
from the impact texture; a guess is not enough. A short clip caption is not
exhaustive, and omitted events are not verified absent. Do not infer an exact
onset inside a labelled clip.

Return one JSON object only. Include each provided window ID exactly once,
in the given order. Use an empty list for an apparently quiet clip.
Do not copy a source guess from one segment into another without hearing it
there.

\begin{verbatim}
{
  "segments": [
    {
      "id": "W0",
      "events": [
        {
          "type": "impact|vocalization|ambience|speech|music|other",
          "heard": "audible texture only",
          "source": "specific source or unresolved",
          "source_confidence": "high|medium|low",
          "contact_material": "identifiable or unresolved"
        }
      ],
      "limitations": []
    }
  ]
}
\end{verbatim}
\end{rewritepromptbox}

\begin{rewritepromptbox}[AAC Silent-video Action Observer]
You are a silent-video action observer. Inspect only this
\texttt{\{duration\_seconds\}}-second video sampled at
\texttt{\{video\_fps\}} fps. The labelled windows are
\texttt{\{window\_boundaries\}}.

For \textbf{each window}, report specific visible actors, actual actions,
motion, pauses, changes, and directly visible contact or other potentially
sound-producing events.

Camera bobbing alone does not prove footsteps. Subject motion together with
scene displacement may support locomotion. Do not imagine sounds.

Report only visible evidence. If a subject leaves the frame or its mouth
is occluded, record the visibility limitation rather than inferring an
associated sound or its absence.

Return JSON only. Return exactly
\texttt{\{number\_of\_windows\}} segments in order, one per supplied window.
Make no audio claims.

\begin{verbatim}
{
  "segments": [
    {
      "id": "W0",
      "time_s": "...",
      "visible_actors": "...",
      "visible_actions": "...",
      "changes": "...",
      "direct_sound_action_evidence": "... or unresolved",
      "limitations": []
    }
  ],
  "scene": "brief setting"
}
\end{verbatim}

\end{rewritepromptbox}

For the reported experiment, \texttt{video\_fps = 2}. The placeholders are
filled separately for each segment of up to 20 seconds, and
\texttt{window\_boundaries} enumerates consecutive 4-second windows
\texttt{W0}, \texttt{W1}, and so on.

\begin{rewritepromptbox}[Action--Audio Correspondence (AAC) Judge]
You are an evidence-based judge of \textbf{Action--Audio Correspondence (AAC)}.
Assign an integer score from 1 to 5 using only the supplied audio-only and
silent-video observations.

The input consists of independent audio-only event observations and silent-video
action observations for the same labelled time windows. These observations are
evidence, not ground truth. Do not fill gaps with imagined sounds or actions.

Assess audible responses to visible actions and their coarse temporal relation
across the full excerpt.

\textbf{Scope.}
First identify concrete visible actions for which an audible response can
reasonably be evaluated, such as foot or object contact, splashing, mechanical
operation, or directly visible vocalization.

Confirmed locomotion may also qualify when subject motion and scene displacement
jointly establish travel, even if feet are occluded; camera bobbing alone is
insufficient.

Test whether a plausible rhythmic or event sound appears, changes, pauses, or
stops in the corresponding windows. A sound can continue while its source is
offscreen. Do not require every visible movement to make sound.

A scene label or mere coexistence of a source and sound does not establish
action--audio correspondence. A flying bird together with bird calls is not
evidence that a visible vocalization caused a call. Generic rushing together
with visible waves is not evidence of a response change.

Do not judge general acoustic fidelity, synthetic artifacts, or scene ambience
without an identifiable action; those are outside AAC.

\textbf{Evidence rules.}
Claim a missing sound only if the action is visible, the sound would reasonably
be audible, absence was explicitly checked, and masking is considered.

Claim a mismatch only from supported observations, not from a sparse caption's
omission.

An animal may vocalize while its mouth is hidden or too small to resolve.
Lack of visible mouth movement is \textbf{not} evidence that an audible call
is detached.

A transition between water, stone, grass, or other materials does not require
an audible timbre change unless contact and the specific sound relation are
independently verified; material matching is outside AAC.

Mechanical repetition, muffling, and other acoustic defects belong to AR and
must not lower AAC by themselves.

Do not infer exact contact timing inside a window, precise source location, or
frame-level synchronization from 2~fps video and window-level audio.
State uncertainty when an audible event's source cannot be identified.
Do not use the prompt's intended action as evidence that the video performed it.

\textbf{Rating.}

\textbf{5:} Multiple supported action/sound changes align, including starts,
stops, pauses, or intensity changes, with no meaningful contradiction.

\textbf{4:} At least one specific supported action/sound change aligns at the
available time resolution, with one minor lapse or limited further variation.

\textbf{3:} At least one specific supported action/sound response exists, but
timing evidence is weak or mixed. Sustained, visually verified locomotion plus
plausible rhythmic contact sounds without observable starts or stops is at most
3, and source uncertainty must be stated.

\textbf{2:} Major supported missing, displaced, or detached responses dominate,
though some relationship remains.

\textbf{1:} Clear, pervasive detachment across the assessable actions and time
windows.

Do not force a target distribution or penalize uncertainty as failure.

To assign a score of 3--5, cite an identifiable visible sound-producing action
and a plausible corresponding audible event. Sustained confirmed locomotion
with plausible rhythmic contacts is allowed but capped at 3 when footfalls are
unseen and timing changes are absent.

To assign a score of 4--5, additionally cite a supported temporal change in
both modalities, with window IDs.

If the evidence consists only of birds and calls coexisting, or generic waves
and rushing coexisting, return \texttt{status="needs\_review"} with no score,
regardless of apparent scene plausibility.

If no action--audio relation is actually assessable, return
\texttt{status="needs\_review"} and omit the score.

Return JSON only. Omit \texttt{score} when
\texttt{status="needs\_review"}. Every cited ID must exist in the supplied
observations.

\begin{verbatim}
{
  "metric": "AAC",
  "status": "scored|needs_review",
  "score": 1,
  "evidence": [
    {
      "window": "W0",
      "audio_ids": ["W0:0"],
      "video_ids": ["W0"],
      "relation": "supported action/sound relation"
    }
  ],
  "reason": "brief evidence-based anchor justification",
  "limitations": []
}
\end{verbatim}
\end{rewritepromptbox}

\subsection{Failure-Type Diagnostic Prompt}
\label{app:failure_type_prompt}

To support qualitative diagnosis beyond aggregate evaluation scores, we use
a vision--language judge to identify recurrent observable failure modes in
long-horizon interactive rollouts. The judge performs multi-label
classification over a predefined failure taxonomy covering temporal
instability, appearance and geometry drift, revisit inconsistency, control
response errors, physical violations, causal-state errors, and failures of
ongoing scene evolution. Multiple failure types may be assigned to the same
video when they correspond to distinct observable phenomena.

The prompt requires the judge to rely only on directly observable video
evidence, avoid inferring hidden model mechanisms, and abstain from assigning
a label when the evidence is weak or ambiguous. It also specifies boundaries
between closely related failure modes, such as global temporal jumps versus
entity-level impossible motion, revisit-content mismatch versus loop-closure
failure, and global first-frame resets versus local state reversion. The
output is restricted to a JSON list containing only labels from the predefined
taxonomy; an empty list indicates that no clearly supported failure is
observed.

\begin{rewritepromptbox}[Long-Horizon Failure-Type Judge]
You are an evidence-based evaluator of failure modes in
\textbf{long-horizon interactive world models}.

Your task is to identify which observable failure types occur in the generated
interactive video.

A video may contain \textbf{ZERO, ONE, or MULTIPLE} failure types.
Select all clearly supported failures, but use only the predefined labels
below. Do not invent, rename, merge, or rewrite labels.

\textbf{Allowed failure labels.}

\begin{verbatim}
[
  "Quality_Degradation",
  "Abrupt_Temporal_Jump",
  "First_Frame_Reset",
  "Temporal_Oscillation",
  "Repeated_Segment_Loop",
  "Appearance_Drift",
  "Geometry_Layout_Drift",
  "Revisit_Mismatch",
  "Repeated_Revisit_Drift",
  "Loop_Closure_Failure",
  "Under_Response",
  "Over_Response",
  "Delayed_Response",
  "Residual_Action",
  "Control_Interference",
  "Subject_Camera_Confusion",
  "Penetration",
  "Floating_Support_Failure",
  "Impossible_Motion",
  "Missing_Consequence",
  "Spurious_Consequence",
  "State_Reversion",
  "Process_Freeze",
  "Process_Restart",
  "Temporal_Reversal"
]
\end{verbatim}

\textbf{Failure definitions.}

\textbf{1. Quality\_Degradation.}
The overall visual quality becomes noticeably worse as the rollout continues.
Typical evidence includes increasing blur or noise, texture corruption,
structural artifacts, loss of fine details, or general visual collapse.
Use this label only for degradation of visual quality, not for identity
changes, scene-layout changes, or world-state errors.

\textbf{2. Abrupt\_Temporal\_Jump.}
The overall scene, viewpoint, or world state changes abruptly between adjacent
moments without a continuous transition explained by the intended interaction
or normal scene evolution. Typical evidence includes sudden scene replacement,
an abrupt viewpoint jump, a large unexplained visual discontinuity, or
globally incompatible neighboring moments. Use this label for a
\emph{global} temporal discontinuity.

\textbf{3. First\_Frame\_Reset.}
The generated world abruptly returns to the initial scene, initial viewpoint,
or a clearly near-initial global state, effectively discarding accumulated
interaction history. Use this label only when the reset is specifically
toward the initial state.

\textbf{4. Temporal\_Oscillation.}
The generated world repeatedly alternates between two or more distinct
previously observed states instead of evolving continuously, e.g.,
$A\!\rightarrow\!B\!\rightarrow\!A\!\rightarrow\!B$ or
$A\!\rightarrow\!B\!\rightarrow\!C\!\rightarrow\!B\!\rightarrow\!C$.
Use this label when distinct states repeatedly switch back and forth.

\textbf{5. Repeated\_Segment\_Loop.}
The same or nearly identical frame sequence is repeatedly regenerated,
forming a short temporal cycle. Typical evidence includes the same few frames
or motion pattern repeatedly restarting. Unlike
\texttt{Temporal\_Oscillation}, this failure repeats nearly the same frames or
short sequence rather than alternating among distinct world states.

\textbf{6. Appearance\_Drift.}
A persistent subject, object, or scene element changes its stable visual
identity or attributes without justification. Examples include changes in
identity, clothing, color, shape, texture, or material. Use this label when
the entity should remain the same but its appearance changes.

\textbf{7. Geometry\_Layout\_Drift.}
The spatial structure of the world changes inconsistently over time.
Examples include roads changing shape, walls or buildings moving, room
layouts changing, landmarks shifting, or previously stable spatial
relationships becoming inconsistent. Use this for geometry and layout
changes rather than appearance changes.

\textbf{8. Revisit\_Mismatch.}
After successfully returning to approximately the same previously observed
location or viewpoint, the world content is inconsistent with the earlier
observation. Objects may disappear or appear, landmarks may change, or
geometry and spatial relations may differ. Use this only when the revisit
pose is approximately recovered but the world content is inconsistent.

\textbf{9. Repeated\_Revisit\_Drift.}
Across multiple revisits to the same place or viewpoint, consistency
progressively deteriorates. Use this only when multiple revisits are
available and the mismatch accumulates across them.

\textbf{10. Loop\_Closure\_Failure.}
After a control sequence that should return the camera or subject to an
earlier pose, the generated result fails to recover the expected position
and/or orientation. The final viewpoint may be spatially offset, have an
incorrect heading, or show expected landmarks from the wrong pose.
Unlike \texttt{Revisit\_Mismatch}, the failure is in recovering the pose
itself rather than preserving world content at a recovered pose.

\textbf{11. Under\_Response.}
The observed translation or rotation is substantially weaker than requested
by the control input. Examples include insufficient forward, backward, or
lateral movement, insufficient camera rotation, or only partial execution of
the requested motion.

\textbf{12. Over\_Response.}
The observed translation or rotation is substantially stronger than requested.
Examples include excessive translation or rotation, or a short control
producing an disproportionately large response.

\textbf{13. Delayed\_Response.}
The intended movement or rotation begins noticeably later than the
corresponding control input. Use this when the control is issued but the
expected response appears only after a clear delay.

\textbf{14. Residual\_Action.}
The effect of a previous control continues after that control has ended.
Examples include continued camera rotation or subject movement after the
corresponding input stops, or a previous action persisting into the next
control segment.

\textbf{15. Control\_Interference.}
Two or more simultaneous or sequential controls incorrectly interfere with
one another. Examples include one component of a combined action being
suppressed, previous and current controls becoming mixed, or translation and
rotation becoming incorrectly entangled. Use this only when the failure is
specifically attributable to interaction between multiple controls.

\textbf{16. Subject\_Camera\_Confusion.}
A subject-directed action incorrectly affects the camera, or a
camera-directed action incorrectly affects the subject. Examples include the
camera moving when only the controlled subject should move, the subject
moving when only the camera should rotate, or the two becoming incorrectly
coupled.

\textbf{17. Penetration.}
The camera, subject, or object passes through a solid surface or obstacle.
Examples include walking through walls, the camera passing through geometry,
objects intersecting solid obstacles, or a subject sinking through the floor.

\textbf{18. Floating\_Support\_Failure.}
A subject or object loses physically plausible support. Examples include a
subject floating above the ground, an unsupported object remaining suspended,
or feet, wheels, or objects visibly detaching from their supporting surface.

\textbf{19. Impossible\_Motion.}
A specific subject or object exhibits physically or temporally impossible
motion while the overall scene does not necessarily reset. Examples include
entity teleportation, discontinuous displacement, implausible acceleration,
an impossible trajectory, or a sudden object-position jump. Unlike
\texttt{Abrupt\_Temporal\_Jump}, this label concerns a specific moving entity
rather than a global scene or viewpoint discontinuity.

\textbf{20. Missing\_Consequence.}
An observable action or physical interaction occurs, but its expected visible
consequence does not. Examples include an object being pushed without
responding, visible contact without an expected effect, or an interaction
that should alter the world but produces no visible change.

\textbf{21. Spurious\_Consequence.}
A world-state change occurs without a corresponding visible action or event.
Examples include an object moving by itself, a door opening without
interaction, or another unexplained state change.

\textbf{22. State\_Reversion.}
A previously established local or semantic world-state change is undone
without a later event that justifies the reversal. Examples include a moved
object returning to its previous location, an opened door becoming closed
again, or an earlier action consequence disappearing. Unlike
\texttt{First\_Frame\_Reset}, this affects a specific state or consequence
rather than globally resetting the world toward its initial state.

\textbf{23. Process\_Freeze.}
An ongoing dynamic process unexpectedly stops. Examples include walking
characters freezing, traffic stopping, moving background elements becoming
static, or another evolving process halting without explanation.

\textbf{24. Process\_Restart.}
An ongoing dynamic process unexpectedly returns to an earlier stage and starts
evolving again from that stage. Examples include a motion sequence restarting
or a dynamic process repeatedly resetting. Use this for dynamic processes,
not for restoration of a static world state.

\textbf{25. Temporal\_Reversal.}
An ongoing process or state evolution moves backward in an implausible manner.
Examples include visible rewinding, a process evolving backward, objects
traversing previously observed intermediate states in reverse, or a completed
transition reversing without a plausible cause.

\textbf{General decision rules.}

Judge only failures that are directly observable in the video.
Do not infer hidden model mechanisms, internal causes, or unobserved events.

Do not treat intended camera motion, normal viewpoint changes, temporary
occlusion, ordinary motion blur, legitimate physical interaction, or
justified scene evolution as failures.

Select \textbf{all} clearly supported failure labels. Multiple labels may be
assigned to the same video when they describe genuinely distinct observable
failures.

Do not assign multiple labels to the same phenomenon merely because several
labels could loosely describe it. Prefer the most specific applicable label.

If the evidence for a failure is weak, ambiguous, or depends on an unsupported
inference, do not select that label.

If no clear failure is visible, return an empty list.

\textbf{Important label boundaries.}

Use \texttt{Quality\_Degradation} when global visual quality becomes worse,
and \texttt{Appearance\_Drift} when a persistent entity changes identity or
visual attributes.

Use \texttt{Geometry\_Layout\_Drift} when stable world geometry or spatial
layout changes.

Use \texttt{Abrupt\_Temporal\_Jump} for a sudden global scene or viewpoint
discontinuity, and \texttt{Impossible\_Motion} for impossible displacement of
a specific entity.

Use \texttt{First\_Frame\_Reset} when the whole generated world abruptly
returns toward its initial state.

Use \texttt{Temporal\_Oscillation} when two or more distinct states repeatedly
alternate, and \texttt{Repeated\_Segment\_Loop} when essentially the same
short frame sequence repeats.

Use \texttt{Revisit\_Mismatch} when a previously observed pose is
approximately recovered but its world content is inconsistent.
Use \texttt{Repeated\_Revisit\_Drift} when such inconsistencies progressively
increase over multiple revisits.
Use \texttt{Loop\_Closure\_Failure} when the expected return pose itself is
not correctly recovered.

Use \texttt{State\_Reversion} when a specific established state or action
consequence is unexpectedly undone.

Use \texttt{Process\_Freeze} when an ongoing process stops,
\texttt{Process\_Restart} when it returns to an earlier stage and begins
again, and \texttt{Temporal\_Reversal} when its evolution visibly proceeds
backward.

\textbf{Output format.}

Return exactly one JSON list. Each item must be exactly one label from the
predefined label set.

Example with multiple failures:
\begin{verbatim}
[
  "First_Frame_Reset",
  "Temporal_Oscillation",
  "Geometry_Layout_Drift",
  "State_Reversion"
]
\end{verbatim}

Example with one failure:
\begin{verbatim}
[
  "Quality_Degradation"
]
\end{verbatim}

Example with no detected failure:
\begin{verbatim}
[]
\end{verbatim}

\textbf{Strict output constraints.}

Output valid JSON only.
Output exactly one JSON list.
Do not output Markdown, explanations, timestamps, confidence scores,
severity scores, category names, or any additional text.
Do not output duplicate labels.
Do not invent or rename labels.
Every output label must exactly match one item in the predefined label set.
\end{rewritepromptbox}
\end{document}